%% file: main.tex
\pdfoutput=1

\documentclass[11pt]{article}

\usepackage[a4paper,
            left=1in,
            right=1in,
            top=1in,
            bottom=1in,
            ]{geometry}

\usepackage{natbib}
 \setcitestyle{aysep={}}	
 \renewcommand{\cite}{\citep}	

\usepackage{times}
\usepackage{latexsym}

\newcommand{\ArgMin}[0]{ArgMin}

\newcommand{\Claim}[0]{C}
\newcommand{\MajorClaim}[0]{MC}
\newcommand{\Premise}[0]{P}
\newcommand{\ImplicitPremise}[0]{IP}
\newcommand{\Background}[0]{B}
\newcommand{\Recommendation}[0]{R}
\newcommand{\Other}[0]{O}

\newcommand{\BARTbase}[0]{\texttt{BART-base}}
\newcommand{\Tbase}[0]{\texttt{T5-base}}
\newcommand{\Tlarge}[0]{\texttt{T5-large}}
\newcommand{\BERT}[0]{\texttt{BERT}}
\newcommand{\XLMr}[0]{\texttt{XLM-R}}
\newcommand{\BARTvone}[0]{\texttt{BART V1}}
\newcommand{\BARTvtwo}[0]{\texttt{BART V2}}
\newcommand{\BERTtwoADUs}[0]{\texttt{BERT} (``2 ADUs'')}
\newcommand{\BERTthreeADUs}[0]{\texttt{BERT} (``3 ADUs'')}
\newcommand{\BERTpredtypeone}[0]{\texttt{BERT} (``pred type 1'')}
\newcommand{\BERTpredtypetwo}[0]{\texttt{BERT} (``pred type 2'')}
\newcommand{\BERTall}[0]{\texttt{BERT} (``all'')}

\usepackage[T1]{fontenc}

\usepackage[utf8]{inputenc}

\usepackage{microtype}

\usepackage{graphicx}
\usepackage{multirow}
\usepackage{hyperref}
\usepackage{xcolor}

\definecolor{darkblue}{rgb}{0, 0, 0.5}

\hypersetup{colorlinks=true,citecolor=darkblue, linkcolor=darkblue, urlcolor=darkblue}

\title{Cross-Genre Argument Mining: Can Language Models \\ Automatically Fill in Missing Discourse Markers?}

\author{
Gil Rocha$^{\dagger}$, Henrique Lopes~Cardoso$^{\dagger}$, Jonas Belouadi$^{\ddagger}$, Steffen Eger$^{\ddagger}$ \\
\small $^{\dagger}$ Laborat\'orio de Intelig\^encia Artificial e Ci\^encia de Computadores (LIACC) \\
\small Faculdade de Engenharia da Universidade do Porto, Porto, Portugal \\
\small $^{\ddagger}$ Natural Language Learning Group (NLLG) \\ 
\small Faculty of Technology, Bielefeld University \\
\small \texttt{\{gil.rocha, hlc\}@fe.up.pt} \\
\small \texttt{\{jonas.belouadi, steffen.eger\}@uni-bielefeld.de}
}
\date{}

\begin{document}
\maketitle
\begin{abstract}
Available corpora for Argument Mining differ along several axes, and one of the key differences is the presence (or absence) of discourse markers to signal argumentative content. 
Exploring effective ways to use discourse markers has received wide attention in various discourse parsing tasks, from which it is well-known that discourse markers are strong indicators of discourse relations. 
To improve the robustness of Argument Mining systems across different genres, we propose to automatically augment a given text with discourse markers such that all relations are explicitly signaled. 
Our analysis unveils that popular language models taken out-of-the-box fail on this task; however, when fine-tuned on a new heterogeneous dataset that we construct (including synthetic and real examples), they perform considerably better. 
We demonstrate the impact of our approach on an Argument Mining downstream task, evaluated on different corpora, showing that language models can be trained to automatically fill in discourse markers across different corpora, improving the performance of a downstream model in some, but not all, cases. 
Our proposed approach can further be employed as an assistive tool for better discourse understanding.  

\end{abstract}

\input{structure/01_introduction}

\input{structure/04_synthetic_data}

\input{structure/03_data}

\input{structure/05_endtoend_augmentation}

\input{structure/06_downstream}

\input{structure/02_related_work}

\input{structure/07_conclusion}

\input{structure/99_limitations}

\section*{Acknowledgments}
Gil Rocha is supported by a PhD grant (SFRH/BD/140125/2018) from Fundação para a Ciência e a Tecnologia (FCT). 
This work was supported by LIACC, funded by national funds through FCT/MCTES (PIDDAC), with reference UIDB/00027/2020. 
The NLLG group is supported by the BMBF grant ``Metrics4NLG'' and the DFG Heisenberg Grant EG 375/5--1.

\bibliography{custom}

\appendix

\input{structure/appendices}

\end{document}

%% file: structure/01_introduction.tex
\section{Introduction}
\label{sec:intro}

%
%
Argument Mining (\ArgMin{}) is a discourse parsing task that aims to automatically extract structured arguments from text. 
In general, an argument in NLP and machine learning is a graph-based structure, where nodes correspond to Argumentative Discourse Units (ADUs), which are connected via argumentative relations (e.g., support or attack)~\citep{LiTo16}. 
%
%
Available corpora for \ArgMin{} differ along several axes, such as language, genre, domain, and annotation schema~\citep{LiTo16,lawrence-reed-2019-argument}. 
One key aspect that differs across different corpora (and even across different articles in the same corpus) is the presence (or absence) of discourse markers (DMs)~\citep{eger-etal-2018-cross,rocha-etal-2018-cross}. 
These DMs are lexical clues that typically precede ADUs. 
%
%
Exploring effective ways to use DMs has received wide attention in various NLP tasks~\citep{pan-etal-2018-discourse,sileo-etal-2019-mining}, including \ArgMin{} related tasks~\citep{kuribayashi-etal-2019-empirical,opitz-frank-2019-dissecting}. 
In discourse parsing~\citep{Mann1987,prasad-etal-2008-penn}, DMs are known to be strong cues for the identification of discourse relations~\citep{Marcu00,braud-denis-2016-learning}. 
Similarly, for \ArgMin{}, the presence of DMs are strong indicators for the identification of ADUs~\citep{eger-etal-2018-cross} and for the overall argument structure \citep{stab-gurevych-2017-parsing, kuribayashi-etal-2019-empirical} (e.g., some DMs are clear indicators of the ADU role, such as premise, conclusion, or major claim).  

%
%
The absence of DMs makes the task more challenging, requiring the system to more deeply capture semantic relations between ADUs~\citep{Moens17}. 
Ideally, an \ArgMin{} system should be able to exploit the presence of explicit DMs as clear indicators of the writer's intention to better capture the argument structure conveyed in the text. 
However, when such surface-level indicators are not provided, the system should be robust enough to capture implicit relations between the corresponding ADUs.

%
%
To close the gap between these two scenarios (i.e.,\ relations explicitly signaled with DMs vs.\ implicit relations), we ask whether recent large language models (LLMs) such as BART~\citep{lewis-etal-2020-bart}, T5~\citep{RaShRo20} and ChatGPT\footnote{\url{https://openai.com/blog/chatgpt/}}, can be used to automatically augment a given text with DMs such that all relations are explicitly signaled.  
Due to the impressive language understanding and generation abilities of recent LLMs, we speculate that such capabilities could be leveraged to automatically augment a given text with DMs. 
However, our analysis unveils that such language models (LMs), when employed in a zero-shot setting, underperform in this task. 
%
%
To overcome this challenge, we hypothesize that a sequence-to-sequence (Seq2Seq) model fine-tuned to augment DMs in an end-to-end setting (from an original to a DM-augmented text) should be able to add coherent and grammatically correct DMs, thus adding crucial signals for \ArgMin{} systems. 
\begin{figure}[t]
	\centerline{\includegraphics[width=0.7\textwidth,keepaspectratio]{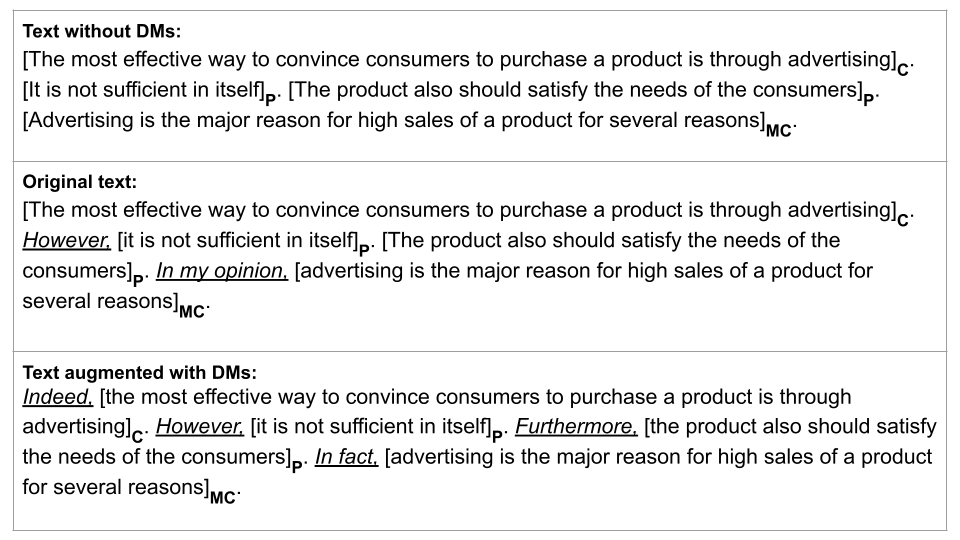}}
	\caption{Illustration of proposed DM augmentation approach. We include annotations for ADU boundaries (square brackets) and labels (acronyms in subscript, see Section~\ref{sec:data} for details regarding ADU labels). 
} 
	\label{fig:intro_example}
\end{figure}
%

%
%
Our second hypothesis is that downstream \ArgMin{} models can profit from automatically added DMs because these contain highly relevant signals for solving \ArgMin{} tasks, such as ADU identification and classification. 
Moreover, given that the Seq2Seq model is fine-tuned on heterogeneous data, we expect it to perform well across different genres. 
To this end, we demonstrate the impact of our approach on an \ArgMin{} downstream task, evaluated on different corpora. 

The proposed approach is illustrated in Figure~\ref{fig:intro_example}.  
The example contains different versions of the same paragraph. 
The version ``Original text'' was extracted from the Persuasive Essays corpus (PEC) \citep{stab-gurevych-2017-parsing} and contains explicit DMs provided by the essay's author. 
``Text without DMs'' shows how challenging it is to grasp the argument structure when the text is deprived of DMs. 
Finally, ``Text augmented with DMs'' illustrates a DM-augmented version of the paragraph based on our proposed approach, where the automatic explicitation of DMs provides useful indicators to unveil the argumentative content. 

Our experiments indicate that the proposed Seq2Seq models can augment a given text with relevant DMs; however, the lack of consistency and variability of augmented DMs impacts the capability of downstream task models to systematically improve the scores across different corpora. 
Overall, we show that the DM augmentation approach improves the performance of \ArgMin{} systems in some corpora and that it provides a viable means to inject explicit indicators when the argumentative reasoning steps are implicit. 
Besides improving the performance of \ArgMin{} systems, especially across different genres, we believe that this approach can be useful as an assistive tool for discourse understanding, e.g., in education contexts. 

In summary, our main contributions are:  
(i) we propose a synthetic template-based test suite, accompanied by automatic evaluation metrics and an annotation study, to assess the capabilities of state-of-the-art LMs to predict DMs;  
(ii) we analyze the capabilities of LMs to augment text with DMs, finding that they underperform in this task;  
(iii) we compile a heterogeneous collection of DM-related datasets on which we fine-tune LMs, 
showing that we can substantially improve their ability for DM augmentation, and 
(iv) we evaluate the impact of end-to-end DM augmentation in a downstream task and find that it improves the performance of \ArgMin{} systems in some, but not all, cases.

%% file: structure/04_synthetic_data.tex
\section{Fill-in-the-mask discourse marker prediction}
\label{sec:fitm}
We now assess whether current state-of-the-art language models can predict coherent and grammatically correct DMs.
To this end, we create an artificial dataset that allows us to evaluate whether language models are sensitive to specific semantically-critical edits in the text. 
These targeted edits are based on the DMs that precede each ADU and the claim's stance. 
When manipulated accordingly, the edits can entail relevant changes in the argument structure. 

\begin{figure}[h]
	\centerline{\includegraphics[width=0.5\textwidth,keepaspectratio]{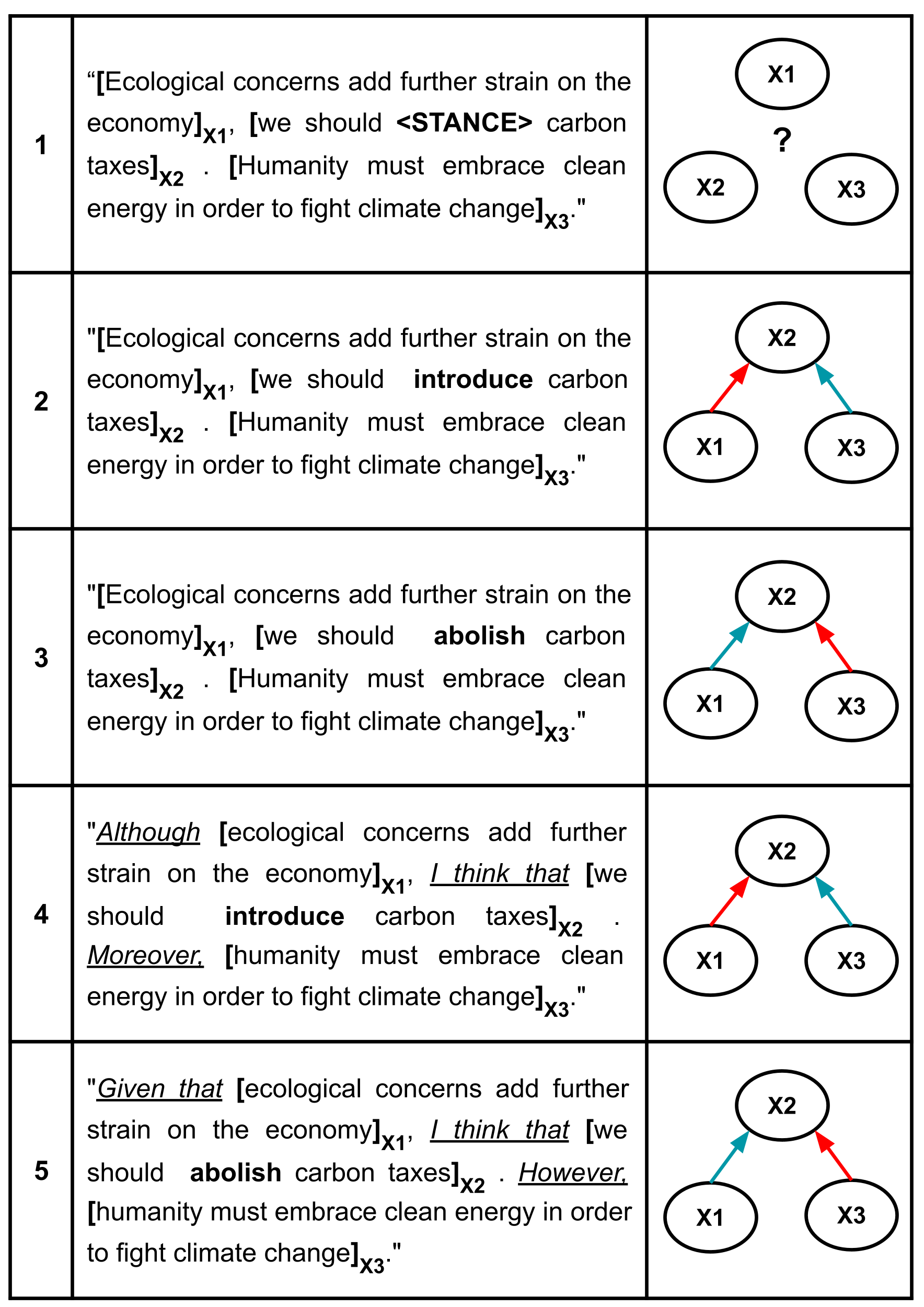}}
	\caption{Artificial dataset example}
	\label{fig:artificial_dataset_example}
\end{figure}

To illustrate the crucial role of DMs and stance for argument perception, consider Example~1 in Figure~\ref{fig:artificial_dataset_example}, in which we have omitted the DMs and the stance-revealing word. 
Without these pieces of information, we are unable to map the argument to a concrete structure. 
Examples~2 and~3 show that it is possible to obtain different text sequences with opposite stances, illustrating the impact that a single word (the stance in this case) can have on the structure of the argument. 
Indeed, the role of the premises changes according to the stance (e.g., ``X1'' is an attacking premise in Example 2 but a supportive premise in Example 3), reinforcing that the semantically-critical edits in the stance impact the argument structure. 
We also note that, even though the text sequences are deprived of DMs, we can infer (based on the semantics of the content) the corresponding argument structure by revealing the stance. 
Finally, Examples 4 and 5 show that making the DMs explicit improves the readability and makes the argument structure clearer (reducing the cognitive interpretation demands required to unlock the argument structure compared to Examples 2 and 3). 
Examples 4 and 5 also show that by changing the stance, adjustments of the DMs that precede each ADU are required to obtain a coherent text sequence. 
Overall, Figure~\ref{fig:artificial_dataset_example} illustrates the key property that motivated our artificial dataset: subtle edits in  content (i.e., stance and DMs) might have a relevant impact on the argument structure. 
On the other hand, some edits do not entail any difference in the argument structure (e.g., the position of ADUs, such as whether the claim occurs before/after the premises). 

To assess the sensitivity of language models to capture these targeted edits, we use a ``fill-in-the-mask'' setup, where some of the DMs are masked, and the models aim to predict the masked content. 
To assess the robustness of language models, we propose a challenging testbed by providing text sequences that share similar ADUs but might entail different argument structures based on the semantically-critical edits in the text previously mentioned (Section~\ref{ssec:artificial_dataset}). 
To master this task, models are not only required to capture the semantics of the sentence (claim's stance and role of each ADU) but also to take into account the DMs that precede the ADUs (as explicit indicators of other ADUs' roles and positioning).

\subsection{Artificial dataset}
\label{ssec:artificial_dataset}
Each sample comprises a claim and one (or two) premise(s) such that we can map each sample to a simple argument structure. 
Every ADU is preceded by one DM that signals its role. 
Claims have a clear stance towards a given position (e.g., ``we should introduce carbon taxes''). 
This stance is identifiable by a keyword in the sentence (e.g., ``introduce''). 
To make the dataset challenging, we also provide samples with opposite stances (e.g., ``introduce'' vs.\ ``abolish''). 
We have one premise in support of a given $\langle claim, stance \rangle$, and another against it. For the opposite stance, the roles of the premises are inverted (i.e., the supportive premise for the claim with the original stance becomes the attacking premise for the claim with the opposite stance). 
For the example in Figure~\ref{fig:artificial_dataset_example}, we have the following core elements:
claim = ``we should <STANCE> carbon taxes'', 
original stance = ``introduce'', 
opposite stance = ``abolish'', 
premise support (original stance) = ``humanity must embrace clean energy in order to fight climate change'', and 
premise attack (original stance) = ``ecological concerns add further strain on the economy''.

Based on these core elements, we follow a template-based procedure to generate different samples. 
The templates are based on a set of configurable parameters: number of ADUs $\in \{2, 3\}$; stance role $\in \{original, opposite\}$; claim position $\in \{1, 2\}$; premise role $\in \{support, attack\}$; supportive premise position $\in \{1, 2\}$; prediction type $\in \{\textit{dm}_{1}, \textit{dm}_{2}, \textit{dm}_{3}\}$. 
Additional details regarding the templates can be found in Appendix~\ref{app:art_dataset_templates}. 
We generate one sample for each possible configuration, resulting in $40$ samples generated for a given instantiation of the aforementioned core elements. 

DMs are added based on the role of the ADU that they precede, using a fixed set of DMs (details provided in Appendix~\ref{app:art_dataset_dms}). 
Even though using a fixed set of DMs comes with a lack of diversity in our examples, the main goal is to add gold DMs consistent with the ADU role they precede. 
We expect that the language models will naturally add some variability in the set of DMs predicted, which we compare with our gold DMs set (using either lexical and semantic-level metrics, as introduced in Section~\ref{ssec:auto_eval_metrics}).  

Examples~4 and~5 in Figure~\ref{fig:artificial_dataset_example} illustrate two samples from the dataset.  
In these examples, all possible masked tokens are revealed (i.e.,\ the underlined DMs). 
The parameter ``prediction type'' will dictate which of these tokens will be masked for a given sample.  

To instantiate the ``claim'', ``original stance'', ``premise support'' and ``premise attack'', we employ the Class of Principled Arguments (CoPAs) set provided by \citet{bilu-etal-2019-argument}. 
Details regarding this procedure can be found in Appendix~\ref{app:art_dataset_copas}. 
 
The test set contains $15$ instantiations of the core elements (all from different CoPAs), resulting in a total of $600$ samples ($40$ samples per instantiation $\times$ $15$ instantiations). 
For the train and dev set, we use only the original stance role (i.e., stance role = $original$), reducing to $20$ the number of samples generated for each core element instantiation. 
We have $251$ and $38$ instantiations of the core elements\footnote{The train set contains $12$ CoPAs, and the dev set contains $2$. To increase the number of instantiations in these partitions, we include all possible motions for each CoPA. Note that different instantiations based on the same CoPA have different claim content but share the same premises (i.e., premises are attached to a given CoPA).}
resulting in a total of $5020$ and $760$ samples (for the train and dev set, respectively). 
To avoid overlap of the core elements across different partitions, a CoPA being used in one partition is not used in another. 

\subsection{Automatic evaluation metrics}
\label{ssec:auto_eval_metrics}

To evaluate the quality of the predictions provided by the models (compared to the gold DMs), we explore the following metrics. 
(1) \emph{Embeddings-based text similarity}: ``word embs'' based on pre-trained embeddings from spaCy library\footnote{\url{https://spacy.io/}}, ``retrofit embs'' based on pre-trained embeddings from LEAR~\citep{vulic-mrksic-2018-specialising}, ``sbert embs'' using pre-trained sentence embeddings from the SBERT library~\citep{reimers-gurevych-2019-sentence}. 
(2) \emph{Argument marker sense} (``arg marker''): DMs are mapped to argument marker senses (i.e., ``forward'', ``backward'', ``thesis'', and ``rebuttal'') \citep{stab-gurevych-2017-parsing}; we consider a prediction correct if the predicted and gold DM senses match.  
(3) \emph{Discourse relation sense} (``disc rel''): we proceed similar to ``arg marker'' but using discourse relation senses~\citep{das-etal-2018-constructing}. These senses are organized hierarchically into 3 levels (e.g., ``Contingency.Cause.Result'') -- in this work, we consider only the first level of the senses (i.e., ``Comparison'', ``Contingency'', ``Expansion'', and ``Temporal''). 
Appendix~\ref{app:auto_eval_metrics} provides additional details on how these metrics are computed. 

For all metrics, scores range from 0 to 1, and obtaining higher scores means  performing better (on the corresponding metric). 

\subsection{Models}
\label{ssec:fitm_models}

We employ the following LMs: 
BERT (``bert-base-cased'') \citep{devlin-etal-2019-bert}, XLM-RoBERTa (XLM-R) (``xlm-roberta-base'') \citep{conneau-etal-2020-unsupervised}, and BART (``facebook/bart-base'')~\citep{lewis-etal-2020-bart}. 
As a first step in our analysis, we frame the DM augmentation problem as a single mask token prediction task. 

For BART, we also report results following a Seq2Seq setting (\BARTvtwo{}), where the model 
receives the same input sequence as the previously mentioned models (text with a single mask token) and 
returns a text sequence as output. 
BART is a Seq2Seq model which uses both an encoder and a decoder from a Transformer-based architecture~\citep{VaShPa17}. Consequently, we can explore the Seq2Seq capabilities of this model to perform end-to-end DM augmentation, contrary to the other models which only contain the encoder component (hence, limited to the single mask-filling setting). 
Including \BARTvtwo{} in this analysis provides a means to compare the performance between these formulations: simpler single-mask prediction from \BARTvone{} vs Seq2Seq prediction from \BARTvtwo{}. 
For the Seq2Seq setting, determining the predicted DM is more laborious, requiring a comparison between the input and output sequence. 
Based on a diff-tool implementation\footnote{\url{https://docs.python.org/3/library/difflib.html}}, we determine the subsequence from the output sequence (in this case, we might have a varied number of tokens being predicted, from zero to several tokens) that matches the ``<mask>'' token in the input sequence. 
Note that the output sequence might contain further differences as compared to the input sequence (i.e., the model might perform further edits to the sequence besides mask-filling); however, these differences are discarded for the fill-in-the-mask DM prediction task evaluation.

\subsection{Experiments}

Table~\ref{tab:fitm_results} shows the results obtained on the test set of the Artificial dataset for the fill-in-the-mask DM prediction task. 

\begin{table}[t]
\centering
\footnotesize
\begin{tabular}{cccccc}
\hline
\multicolumn{1}{c|}{} & \begin{tabular}[c]{@{}c@{}}word \\ embs\end{tabular} & \begin{tabular}[c]{@{}c@{}}retrofit \\ embs\end{tabular} & \begin{tabular}[c]{@{}c@{}}sbert \\ embs\end{tabular} & \begin{tabular}[c]{@{}c@{}}arg \\ marker\end{tabular} & \begin{tabular}[c]{@{}c@{}}disc \\ rel\end{tabular} \\ \hline \hline
\multicolumn{6}{l}{\textit{Zero-shot}} \\ \hline
\multicolumn{1}{c|}{BERT} & \textbf{.7567} & \textbf{.6336} & \textbf{.4605} & .2444 & \textbf{.4361} \\
\multicolumn{1}{c|}{XLM-R} & .7196 & .6080 & .4460 & \textbf{.2500} & .3944 \\
\multicolumn{1}{c|}{BART V1} & .6225 & .4599 & .3165 & .1861 & .3194 \\
\multicolumn{1}{c|}{BART V2} & .6934 & .4325 & .3093 & .0833 & .2417 \\ \hline \hline
\multicolumn{6}{l}{\textit{Fine-tuned (BERT)}} \\ \hline
\multicolumn{1}{c|}{2 ADUs} & .7382 & .5981 & .5659 & .4528 & .6083 \\
\multicolumn{1}{c|}{3 ADUs} & .9687 & .9296 & .8688 & .8333 & .8333 \\
\multicolumn{1}{c|}{pred type 1} & .6886 & .6024 & .5879 & .5278 & .5417 \\
\multicolumn{1}{c|}{pred type 2} & .6935 & .6080 & .5944 & .5361 & .5361 \\
\multicolumn{1}{c|}{all} & \textbf{.9786} & \textbf{.9519} & \textbf{.9103} & \textbf{.8861} & \textbf{.8861} \\ \hline
\end{tabular}
\caption{Fill-in-the-mask DM predictions results. Evaluated on the test set of the Artificial dataset.}
\label{tab:fitm_results}
\end{table}


In a zero-shot setting,  
\BERT{} performs clearly best, leading all evaluation dimensions except for ``arg marker''. 
BART models obtain the lowest scores for all metrics. 

We explore fine-tuning the models on the Artificial dataset (using the train and dev set partitions). 
We use BERT for these fine-tuning experiments. 
Comparing fine-tuned \BERTall{} with zero-shot \BERT{}, we observe clear improvements in all metrics. 
Increases are particularly striking for ``arg marker'' and ``disc rel'' where models improve from 0.24-0.44 to almost 0.9. 
Thus, we conclude that the DM slot-filling task is very challenging in a zero-shot setting for current LMs, but after fine-tuning, the performance can be clearly improved. 

To analyze the extent to which models overfit the fine-tuning data, we selected samples from the training set by constraining some of the parameters used to generate the artificial dataset. 
The test set remains the same as in previous experiments. 
In \BERTtwoADUs{}, only samples containing a single sentence are included (besides single sentences, the test set also includes samples containing two sentences with 3 ADUs that the model was not trained with). 
For \BERTthreeADUs{}, only samples containing two sentences with 3 ADUs are included 
(thus, even though the model is trained to make predictions in the first sentence, it is not explicitly trained to make predictions when only a single sentence is provided, as observed in some samples in the test set).  
In \BERTpredtypeone{}, we only include samples where the mask token is placed at the beginning of the first sentence (consequently, the model is not trained to predict DMs in other positions). 
For \BERTpredtypetwo{}, we only include samples where the mask is placed in the DM preceding the second ADU of the first sentence. 
For both \BERTpredtypeone{} and \BERTpredtypetwo{}, the model is not trained to make predictions in the second sentence; hence, we can analyze whether models generalize well in these cases. 

Comparing the different settings explored for fine-tuning experiments, we observe that constraining the training data to single-sentence samples (``2 ADUs'') or to a specific ``pred type'' negatively impacts the scores (performing even below the zero-shot baseline in some metrics). 
These findings indicate that the models fail to generalize when the test set contains samples following templates not provided in the training data. 
This exposes some limitations of recent LMs.

\subsection{Error analysis}

We focus our error analysis on the discourse-level senses: ``arg marker'' and ``disc rel''. 
The goal is to understand if the predicted DMs are aligned (regarding positioning) with the gold DMs. 

Regarding zero-shot models, we make the following observations.  
\BERT{} and \XLMr{} often predict DMs found in the DMs list for both senses, meaning they can predict conventional and sound DMs. However, there is some confusion between the labels ``backward'' and ``forward'' vs.\ ``rebuttal'', and ``Comparison'' vs.\  ``Expansion'', indicating that the models are not robust to challenging ``edge cases'' in the Artificial dataset (i.e.,\ different text sequences in the dataset where subtle changes in the content entail different argument structures and/or sets of DMs). 
BART-based models predict more diverse DMs, increasing the number of predictions not found in the DM lists. 
We observe less confusion between these labels, indicating that these models are more robust to the edge cases.  

As the automatic evaluation metrics indicate, fine-tuned \BERTall{} performs better than the zero-shot models. 
Nonetheless, we still observe some confusion between the labels ``backward'' and ``forward'' vs. ``rebuttal'' and ``Contingency'' vs. ``Comparison'', even though these confusions are much less frequent than in zero-shot models. 

Some examples from the test set of the Artificial Dataset are shown in Table~\ref{tab:fitm_edge_cases_gold} (Appendix~\ref{app:fitm_error_analysis_details}). We also show the predictions made by the zero-shot models and fine-tuned \BERTall{} model in Table~\ref{tab:fitm_edge_cases_preds} (Appendix~\ref{app:fitm_error_analysis_details}).

\subsection{Human evaluation}

We conduct a human evaluation experiment to assess the quality of the predictions provided by the models in a zero-shot setting. 
Furthermore, we also aim to determine if the automatic evaluation metrics correlate with human assessments. 

We selected 20 samples from the Artificial dataset, covering different templates. 
For each sample, we get the predictions provided by each model analyzed in a zero-shot setting (i.e., \BERT{}, \XLMr{}, \BARTvone{}, \BARTvtwo{}), resulting in a total of 80 samples; after removing repeated instances with the same prediction, each annotator analyzed 67 samples. 
Three annotators performed this assessment. 
Each annotator rates the prediction made by the model based on two criteria: 
\begin{itemize}
    \item Grammaticality: \textit{Is the predicted content grammatically correct given the context?} The annotator is asked to rate with one of the following options: (-1): no, incorrect or out-of-context; (0): neutral or ambiguous; (+1): yes, it is appropriate.
    \item Coherence: \textit{Is the connotation of the predicted content correct/appropriate taking into account the surrounding context?} Options: (-1): incorrect, predicted content is in the opposite sense; (0): neutral or ambiguous; (+1): correct, right sense.
\end{itemize}
Appendix~\ref{app:human_eval_instances} shows some of the samples presented to the annotators and corresponding ratings. 

We report inter-annotator agreement (IAA) using Cohen's $\kappa$ metric~\citep{davies1982measuring}, based on the scikit-learn~\citep{PeVaGr11} implementation. 
For ``Grammaticality'', we obtain a Cohen's $\kappa$ score of $0.7543$ (which corresponds to ``substantial'' agreement according to the widely used scale of values proposed by \citet{LaKo77}); for ``Coherence'', a ``moderate'' agreement of $0.5848$.  
Overall, IAA scores indicate that human annotators can perform this task with reasonable agreement; however, assessing the ``Coherence'' criteria consistently (especially considering all the subtle changes in content that lead to different argument structures) is a challenging task that requires more cognitive effort. 
Analyzing disagreements, we observe that most of them occur when one of the annotators provides the label ``neutral or ambiguous'' (0) while the other annotator considers either (+1) or (-1). 
We also perform a correlation analysis between criteria ``Grammaticality'' and ``Coherence'' to determine if the labels provided by each annotator for these criteria are correlated. We obtain Pearson correlation coefficients of $0.0573$, $0.0845$, and $0.1945$ (very low correlation). 
Therefore, we conclude that annotators can use the criteria independently.

To determine whether automatic evaluation metrics (Section~\ref{ssec:auto_eval_metrics}) are aligned with human assessments, we perform a correlation analysis. 
To obtain a gold standard label for each text sequence used in the human evaluation, we average the labels provided by the three annotators. 
The results for the correlation analysis are presented in Table~\ref{tab:corr_analysis_human_and_auto_metrics} (using the Pearson correlation coefficient). 
For the ``Grammaticality'' criterion, ``retrofit embs'' is the metric most correlated with human assessments, followed closely by ``word embs''. 
Regarding the ``Coherence'' criterion, ``disc rel'' is the most correlated metric. 
Intuitively, metrics based on discourse-level senses are more correlated with the ``Coherence'' criterion because they capture the discourse-level role that these words have in the sentence. 

\begin{table}[t]
\centering
\footnotesize
\begin{tabular}{c|cc}
\hline
metric & Grammaticality & Coherence \\ \hline \hline
word embs & .4757 & .2795 \\
retrofit embs & \textbf{.4907} & .3243 \\
sbert embs & .2888 & .3519 \\
arg marker & .2075 & .3757 \\
disc rel & -.0406 & \textbf{.5421} \\ \hline
\end{tabular}
\caption{Correlation analysis between human assessment and automatic evaluation metrics}
\label{tab:corr_analysis_human_and_auto_metrics}
\end{table}

%% file: structure/03_data.tex
\section{Data}
\label{sec:data}

In this section, we introduce the corpora used in the end-to-end DM augmentation (Section~\ref{sec:e2e_dm_augment}) and downstream task (Section~\ref{sec:downstream_task}) experiments. 
In Section~\ref{ssec:data_argmin}, we describe three corpora containing annotations of argumentative content that are used in our experiments only for evaluation purposes. 
Then, in Section~\ref{ssec:data_dm_annotations}, we describe the corpora containing annotations of DMs. 
These corpora are used in our experiments to train Seq2Seq models to perform end-to-end DM augmentation.

\subsection{Argument Mining data}
\label{ssec:data_argmin}

\paragraph{Persuasive Essays corpus (PEC)} \citep{stab-gurevych-2017-parsing} 
contains token-level annotations of ADUs and their relations from student essays. An argument consists of a claim and one (or more) premise(s), connected via argumentative relations: ``Support'' or ``Attack''. 
Arguments are constrained to the paragraph boundaries (i.e., the corresponding ADUs must occur in the same paragraph). 
Each ADU is labeled with one of the following: ``Premise'' (\Premise{}), ``Claim'' (\Claim{}), or ``Major Claim'' (\MajorClaim{}). 
A paragraph might contain zero or more arguments. 
The distribution of token-level labels is: \Claim{} ($15\%$), \Premise{} ($45\%$), \MajorClaim{} ($8\%$), \Other{} ($32\%$). 
PEC contains $402$ essays, $1833$ paragraphs, $1130$ arguments, $6089$ ADUs, and an average of $366$ tokens per essay. 
PEC is a well-established and one of the most widely explored corpora for \ArgMin{} tasks.

\paragraph{Microtext corpus (MTX)} \citep{peldszus2015annotated} 
contains token-level annotations of ADUs from short texts (six or fewer sentences) written in response to trigger questions, such as ``Should one do X''.  
Each microtext consists of one claim and several premises. 
Each ADU is labeled with one of the following: \Premise{} or \Claim{}. 
Note that all tokens are associated with an ADU (MTX does not contain \Other{} tokens). 
It contains $112$ microtexts. 
The distribution of token-level labels is: \Claim{} ($18\%$) and \Premise{} ($82\%$).

\paragraph{Hotel reviews corpus (Hotel)} \citep{liu-etal-2017-using,GaWaZh17} 
contains token-level annotations of ADUs from Tripadvisor hotel reviews. 
Each sub-sentence in the review is considered a clause. Annotators were asked to annotate each clause with one of the following labels: ``Background'' (\Background{}), \Claim{}, ``Implicit Premise'' (\ImplicitPremise{}), \MajorClaim{}, \Premise{}, ``Recommendation'' (\Recommendation{}), or ``Non-argumentative'' (O). 
It contains $194$ reviews, with an average of $185$ tokens per review. 
The distribution of token-level labels is: \Background{} ($7\%$), \Claim{} ($39\%$), \ImplicitPremise{} ($8\%$), \MajorClaim{} ($7\%$), \Premise{} ($22\%$), \Recommendation{} ($5\%$), \Other{} ($12\%$). 
We expect this to be a challenging corpus for ADU boundary detection and classification because: 
(a) it contains user-generated text with several abbreviations and grammatical errors; 
(b) the label space is larger; and 
(c) the original text is mostly deprived of explicit DMs. 

\subsection{Data with DM annotations}
\label{ssec:data_dm_annotations}

\paragraph{Artificial dataset (AD)} 
Dataset proposed in this paper to assess the capabilities of LMs to predict DMs. 
Each sample contains a claim and one (or two) premise(s). 
Every ADU is preceded by one DM that signals its role (further details are provided in Section~\ref{ssec:artificial_dataset}). 

\paragraph{Discovery} \citep{sileo-etal-2019-mining} 
provides a collection of adjacent sentence pairs $\langle s1, s2 \rangle$ and corresponding DM $y$ that occur at the beginning of $s2$.   
This corpus was designed for the Discourse Connective Prediction (DCP) task, where the goal is to determine $y$ (from a fixed set of possible DMs) based on $s1$ and $s2$; e.g., $s1$ = ``\emph{The analysis results suggest that the HCI can identify incipient fan bearing failures and describe the bearing degradation process.}'', $s2$ = ``\emph{The work presented in this paper provides a promising method for fan bearing health evaluation and prognosis.}'', and $y$ = ``\emph{overall,}''. 
Input sequences are extracted from the Depcc web corpus~\citep{panchenko-etal-2018-building}, which consists of English texts collected from commoncrawl web data. 
This corpus differs from related work by the diversity of the DMs collected (a total of 174 different classes of DMs were collected, while related work collected 15 or fewer classes of DMs, e.g., the DisSent corpus~\citep{nie-etal-2019-dissent}). 

\paragraph{PDTB-3} \citep{webber-etal-2016-discourse}\footnote{\url{https://catalog.ldc.upenn.edu/LDC2019T05}} 
contains annotations of discourse relations for articles extracted from the Wall Street Journal (WSJ). These discourse relations describe the relationship between two discourse units (e.g., propositions or sentences) and are grounded on explicit DMs occurring in the text (explicit discourse relation) or in the adjacency of the discourse units (implicit discourse relation). For explicit relations, annotators were asked to annotate: the connective, the two text spans that hold the relation, and the sense it conveys based on the PDTB senses hierarchy. For implicit relations, annotators were asked to provide an explicit connective that best expresses the sense of the relation. This resource has been widely used for research related to discourse parsing. 

%% file: structure/05_endtoend_augmentation.tex
\section{End-to-end discourse marker augmentation}
\label{sec:e2e_dm_augment}

In Section~\ref{sec:fitm}, based on experiments in a challenging synthetic testbed proposed in this work (the Artificial Dataset), we have shown that DMs play a crucial role in argument perception and that recent LMs struggle with DM prediction.  
We now move to more real-world experiments. 
As detailed in Section~\ref{sec:intro}, our goal is to automatically augment a text with DMs such that downstream task models can take advantage of explicit signals automatically added. 
To this end, we need to conceive models that can automatically (a) identify where DMs should be added and (b) determine which DMs should be added based on the context. 
We frame this task as a Seq2Seq problem. 
As input, the models receive a (grammatically sound) text sequence that might (or not) contain DMs. 
The output is the text sequence populated with DMs. 
For instance, for \ArgMin{} tasks, we expect that the DMs should be added preceding ADUs. 

\paragraph{Model}
We employ recent Seq2Seq language models, namely: 
BART (``facebook/bart-base'')~\citep{lewis-etal-2020-bart} and T5 (``t5-base'' and ``t5-large'')~\citep{RaShRo20}. 
We use default ``AutoModelForSeq2SeqLM'' and ``Seq2SeqTrainingArguments'' parameters provided by the HuggingFace library, except for the following: scheduler = ``constant'' and max training epochs = $5$. 

\subsection{Evaluation}
\label{ssec:e2e_dm_augment_eval_protocol} 
As previously indicated, explicit DMs are known to be strong indicators of argumentative content, but whether to include explicit DMs inside argument boundaries is an annotation guideline detail that differs across \ArgMin{} corpora. 
Furthermore, in the original \ArgMin{} corpora (e.g., the corpora explored in this work, Section~\ref{ssec:data_argmin}), DMs are not directly annotated. 
We identify the gold DMs based on a heuristic approach proposed by \citet{kuribayashi-etal-2019-empirical}. 
For PEC, we consider as gold DM the span of text that precedes the corresponding ADU; more concretely, the span to the left of the ADU until the beginning of the sentence or the end of a preceding ADU is reached. 
For MTX, DMs are included inside ADU boundaries; in this case, if an ADU begins with a span of text specified in a DM list,\footnote{The longest occurring one, using the same list proposed by \citet{kuribayashi-etal-2019-empirical}, which is composed of DMs that can be found in PEC and PDTB.} we consider that span of text as the gold DM and the following tokens as the ADU. 
For Hotel, not explored by \citet{kuribayashi-etal-2019-empirical}, we proceed similarly to MTX. 
We would like to emphasize that following this heuristic-based approach to decouple DMs from ADUs (in the case of MTX and Hotel datasets) keeps sound and valid the assumption that DMs typically precede the ADUs; this is already considered and studied in prior work~\citep{kuribayashi-etal-2019-empirical,bao-etal-2021-neural}, only requiring some additional pre-processing steps to be performed in this stage to normalize the \ArgMin{} corpora in this axis. 

To detokenize the token sequences provided by the \ArgMin{} corpora, we use sacremoses\footnote{\url{https://github.com/alvations/sacremoses}}. 
As Seq2Seq models output a text sequence, we need to determine the DMs that were augmented in the text based on the corresponding output sequence. 
Similar to the approach detailed in Section~\ref{ssec:fitm_models}, we use a diff-tool implementation, but in this case, we might have multiple mask tokens (one for each ADU). 
Based on this procedure, we obtain the list of DMs predicted by the model, which we can compare to the list of gold DMs extracted from the original input sequence. 

To illustrate this procedure, consider the example in Figure~\ref{fig:intro_example}. 
The gold DMs for this input sequence are: [``'', ``However'', ``'', ``In my opinion''].  
An empty string means that we have an implicit DM (i.e.,\ no DM preceding the ADU in the original input sequence); for the remaining, an explicit DM was identified in the original input sequence. 
The predicted DMs are: [``Indeed'', ``However'', ``Furthermore'', ``In fact'']. 

In terms of evaluation protocol, we follow two procedures: 
(a) \textbf{Explicit DMs accuracy analysis}: based on the gold explicit DMs in the original input sequence, we determine whether the model predicted a DM in the same location and whether the prediction is correct (using the automatic evaluation metrics described in Section~\ref{ssec:auto_eval_metrics}). For sense-based metrics, only gold DMs that can be mapped to some sense are evaluated. With this analysis, we aim to determine the quality of the predicted DMs (i.e., if they are aligned with gold DMs at the lexical and semantic-level). 
(b) \textbf{Coverage analysis}: based on all candidate DMs (explicit and implicit) that could be added to the input sequence (all elements in the gold DM list), we determine the percentage of DMs that are predicted. The aim of this analysis is to determine to which extent the model is augmenting the data with DMs in the correct locations (including implicit DMs, which could not be evaluated in (a)). 
 
For an input sequence, there may be multiple DMs to add; our metrics average over all occurrences of DMs. 
Then, we average over all input sequences to obtain the final scores, as reported in Table~\ref{tab:e2e_dm_augment_results_pec} (for explicit DMs accuracy analysis) and Table~\ref{tab:e2e_dm_augment_pct_gold_dms_pred} (for coverage analysis). 

Importantly, further changes to the input sequence might be performed by the Seq2Seq model (i.e.,\ besides adding the DMs, the model might also commit/fix grammatical errors, capitalization, etc.), but we ignore these differences for the end-to-end DM augmentation assessment.  

\subsection{Data preparation}
\label{ssec:e2e_dm_augment_training_data}

In the following, we describe the data preparation for each corpus (we use the corpora mentioned in Section~\ref{ssec:data_dm_annotations}) to obtain the input and output sequences (gold data) for training the Seq2Seq models in the end-to-end DM augmentation task. 

\paragraph{Artificial dataset} 
For each sample, we generate a text sequence without any DM (as shown with text sequences ``2'' and ``3'' in Figure~\ref{fig:artificial_dataset_example}) for the input sequence and another text sequence with DMs preceding all ADUs (text sequences ``4'' and ``5'' in Figure~\ref{fig:artificial_dataset_example}) for the output sequence. 
\paragraph{Discovery} 
For each sample, we employ the concatenation of the sentences without any DM in between for the input sequence (i.e.,\ ``s1. s2'') and with the corresponding DM at the beginning of $s2$ for the output sequence (i.e.,\  ``s1. y s2'').  
\paragraph{PDTB-3} 
As input, we provide a version of the original text where all explicit DMs are removed. 
We also perform the following operations to obtain a grammatically sound text: 
(a) if the DM is not at the beginning of a sentence and if it is not preceded by any punctuation mark, we replace the DM with a comma -- other options would be possible in specific cases, but we found the comma substitution to be a reasonable option 
(e.g.,  
``\emph{[...] this is a pleasant rally \textbf{but} it's very selective [...]}'' is converted to ``\emph{[...] this is a pleasant rally, it's very selective [...]}''); 
(b) if the DM occurs at the beginning of a sentence, we uppercase the content that follows immediately after the removed DM. 
As output, we provide a version of the original text where the implicit DMs are also added. 
Adding implicit DMs also requires an extra pre-processing step, namely: if the DM occurs at the beginning of a sentence, we lowercase the content that follows the added DM.

\subsection{Results}
\label{ssec:e2e_dm_augment_results}

\paragraph{Setup} 
For evaluation purposes, we analyze the capability of Seq2Seq models to augment a given text with DMs using two versions of the input data: 
(a) the original input sequence (``Input data: original''), which contains the explicit DMs originally included in the text; 
(b) the input sequence obtained after the removal of all explicit DMs (``Input data: removed DMs''). 
The first setting can be seen as the standard application scenario of our proposed approach, where we ask the Seq2Seq model to augment a given text, which might or might not contain explicit DMs. We expect the model to take advantage of explicit DMs to better capture the meaning of the text and to automatically augment the text with implicit DMs. 
The second setting is challenging because the Seq2Seq model is asked to augment a text deprived of explicit signals. 
To remove the explicit DMs from the original input sequence (``Input data: removed DMs''), we use the annotations of ADUs provided in \ArgMin{} corpora.   
As described in Section~\ref{ssec:e2e_dm_augment_eval_protocol}, we follow a heuristic approach~\citep{kuribayashi-etal-2019-empirical} to identify the gold DMs that precede each ADU. Then, we remove the corresponding DMs from the input sequence and perform the same operations described in Section~\ref{ssec:e2e_dm_augment_training_data} for PDTB-3 to obtain a grammatically sound text sequence (e.g.,  
``\emph{A great number of plants and animals died out \textbf{because} they were unable to fit into the new environment.}'' is converted to ``\emph{A great number of plants and animals died out, they were unable to fit into the new environment}.'').

\paragraph{Explicit DMs accuracy analysis}
Table~\ref{tab:e2e_dm_augment_results_pec} details the results. 
The evaluation is performed on the test set of PEC, employing the automatic evaluation metrics described in Section~\ref{ssec:auto_eval_metrics}. 
We do not employ the remaining corpora from Section~\ref{ssec:data_argmin} because the percentage of explicit DMs is relatively low.

\begin{table}[t]
\centering
\footnotesize
\begin{tabular}{ccccccc}
\hline
\multicolumn{1}{c|}{\begin{tabular}[c]{@{}c@{}}seq2seq\\ model\end{tabular}} & \multicolumn{1}{c|}{\begin{tabular}[c]{@{}c@{}}fine-tune \\ data\end{tabular}} & \begin{tabular}[c]{@{}c@{}}word \\ embs\end{tabular} & \begin{tabular}[c]{@{}c@{}}retrofit \\ embs\end{tabular} & \begin{tabular}[c]{@{}c@{}}sbert \\ embs\end{tabular} & \begin{tabular}[c]{@{}c@{}}arg \\ marker\end{tabular} & \begin{tabular}[c]{@{}c@{}}disc \\ rel\end{tabular} \\ \hline \hline
\multicolumn{7}{l}{\textit{Input data: removed DMs}} \\ \hline
\multicolumn{1}{c|}{\multirow{4}{*}{BART-base}} & \multicolumn{1}{c|}{none} & .0075 & .0029 & .0021 & 0 & 0 \\
\multicolumn{1}{c|}{} & \multicolumn{1}{c|}{Discovery} & .4972 & .1969 & .2416 & .3572 & .2231 \\
\multicolumn{1}{c|}{} & \multicolumn{1}{c|}{AD} & \textbf{.5878} & \textbf{.2619} & .2775 & .2475 & .2596 \\
\multicolumn{1}{c|}{} & \multicolumn{1}{c|}{PDTB} & .3805 & .2035 & .1871 & .2070 & .3236 \\ \hline
\multicolumn{1}{c|}{BART-base} & \multicolumn{1}{c|}{\multirow{3}{*}{\begin{tabular}[c]{@{}c@{}}Discovery + \\ AD + PDTB\end{tabular}}} & .5038 & .2101 & .2563 & .3434 & .2150 \\
\multicolumn{1}{c|}{T5-base} & \multicolumn{1}{c|}{} & .5087 & .2372 & \textbf{.2817} & .4290 & \textbf{.3393} \\
\multicolumn{1}{c|}{T5-large} & \multicolumn{1}{c|}{} & .4992 & .2308 & .2768 & \textbf{.4405} & .3204 \\ \hline \hline
\multicolumn{7}{l}{\textit{Input data: original}} \\ \hline
\multicolumn{1}{c|}{BART-base} & \multicolumn{1}{c|}{\multirow{3}{*}{\begin{tabular}[c]{@{}c@{}}Discovery + \\ AD + PDTB\end{tabular}}} & .9075 & .8871 & .7998 & .1992 & .2383 \\
\multicolumn{1}{c|}{T5-base} & \multicolumn{1}{c|}{} & \textbf{.9125} & \textbf{.8908} & .8236 & .4835 & .5127 \\
\multicolumn{1}{c|}{T5-large} & \multicolumn{1}{c|}{} & .8984 & .8781 & \textbf{.8275} & \textbf{.5745} & \textbf{.6247} \\ \hline
\end{tabular}
\caption{End-to-end DM augmentation results - explicit DMs accuracy analysis}
\label{tab:e2e_dm_augment_results_pec}
\end{table}


We start our analysis with the ``Input data: removed DMs'' setting. 
%
%
First, we observe that the pre-trained \BARTbase{} model underperforms in the DM augmentation task in a zero-shot setting (``none'' in the column ``fine-tune data'') because it will not automatically augment the text with DMs without being explicitly trained to do it. 
%
%
Then, we compare the scores obtained when fine-tuning the pre-trained \BARTbase{} model on each corpus individually (``Discovery'', ``AD'', and ``PDTB''). 
We observe that the best scores on the: 
(a) embeddings-based metrics (i.e., ``word embs'', ``retrofit embs'', and ``sbert embs'') are obtained when fine-tuning \BARTbase{} on AD, which we attribute to the restricted set of DMs used in the training data and, consequently, the predictions made by the model are more controlled towards well-known DMs; 
(b) ``disc rel'' metric is obtained fine-tuning on PDTB, which indicates that this corpus is relevant to improve the models on the ``Coherence'' axis; 
(c) ``arg marker'' metric is obtained fine-tuning on Discovery. 
%
%
We also provide results when fine-tuning on the combination of the three corpora in the training set; 
we consider \BARTbase{}, \Tbase{}, and \Tlarge{} pre-trained models. 
We make the following observations: 
(i) surprisingly, \BARTbase{} performs worse when fine-tuned on the combination of all datasets compared to the best individual results; 
(ii) \Tbase{} is superior to \BARTbase{} in all metrics;  
(iii) \Tbase{} performs better than \Tlarge{} in most metrics (except for ``arg marker''), indicating that larger models do not necessarily perform better in this task. 

%
%
Regarding the ``Input data: original'' setting, we analyze the results obtained after fine-tuning on the combination of the three corpora.  
As expected, we obtain higher scores across all metrics compared to ``Input data: removed DMs'', as the Seq2Seq model can explore explicit DMs (given that we frame it as a Seq2Seq problem, the model might keep, edit, or remove explicit DMs) to capture the semantics of text and use this information to improve on implicit DM augmentation. 
\BARTbase{} underperforms in this setting compared to the T5 models. We observe higher variations in the scores for the metrics ``arg marker'' and ``disc rel'', with \Tlarge{} obtaining remarkable improvements, almost 10 percentage points above \Tbase{}, which itself is 30 points above \BARTbase{}.  

\paragraph{Coverage analysis}
Detailed results are provided in Table~\ref{tab:e2e_dm_augment_pct_gold_dms_pred}. 
The evaluation is performed on the test set of each \ArgMin{} corpus described in Section~\ref{ssec:data_argmin}. 
For reference, the results obtained in the original data (i.e., containing only the original explicit DMs, which corresponds to ``Input data: original'' without the intervention of any Seq2Seq model) are: $73\%$ for PEC, $44\%$ for MTX, and $15\%$ for Hotel. 
We explore the same input data settings and Seq2Seq pre-trained models, and fine-tune with data previously detailed for the explicit DM accuracy analysis. 

\begin{table}[ht]
\centering
\footnotesize
\begin{tabular}{ccccc}
\hline
\multicolumn{1}{c|}{\begin{tabular}[c]{@{}c@{}}seq2seq\\ model\end{tabular}} & \multicolumn{1}{c|}{\begin{tabular}[c]{@{}c@{}}fine-tune\\ data\end{tabular}} & PEC & MTX & Hotel \\ \hline \hline
\multicolumn{5}{l}{\textit{Input data: removed DMs}} \\ \hline
\multicolumn{1}{c|}{\multirow{4}{*}{BART-base}} & \multicolumn{1}{c|}{none} & 2.00 & 0 & 5.71 \\
\multicolumn{1}{c|}{} & \multicolumn{1}{c|}{Discovery} & 88.01 & 80.36 & 32.60 \\
\multicolumn{1}{c|}{} & \multicolumn{1}{c|}{AD} & 88.61 & \textbf{80.51} & 40.21 \\
\multicolumn{1}{c|}{} & \multicolumn{1}{c|}{PDTB} & 61.61 & 54.93 & 22.69 \\ \hline
\multicolumn{1}{c|}{BART-base} & \multicolumn{1}{c|}{\multirow{3}{*}{\begin{tabular}[c]{@{}c@{}}Discovery + \\ AD + PDTB\end{tabular}}} & \textbf{88.72} & 79.42 & \textbf{63.28} \\
\multicolumn{1}{c|}{T5-base} & \multicolumn{1}{c|}{} & 86.92 & 80.07 & 41.91 \\
\multicolumn{1}{c|}{T5-large} & \multicolumn{1}{c|}{} & 85.84 & 73.99 & 28.53 \\ \hline \hline
\multicolumn{5}{l}{\textit{Input data: original}} \\ \hline
\multicolumn{1}{c|}{BART-base} & \multicolumn{1}{c|}{\multirow{3}{*}{\begin{tabular}[c]{@{}c@{}}Discovery + \\ AD + PDTB\end{tabular}}} & \textbf{97.93} & \textbf{94.49} & \textbf{69.40} \\
\multicolumn{1}{c|}{T5-base} & \multicolumn{1}{c|}{} & 95.61 & 92.25 & 45.71 \\
\multicolumn{1}{c|}{T5-large} & \multicolumn{1}{c|}{} & 95.24 & 88.55 & 36.44 \\ \hline
\end{tabular}
\caption{End-to-end DM augmentation results - coverage analysis}
\label{tab:e2e_dm_augment_pct_gold_dms_pred}
\end{table}


%
%
Analyzing the results obtained with the ``Input data: removed DMs'' setting, we observe that: \BARTbase{} employed in a zero-shot setting underperforms the task (because it will not augment the text with DMs); fine-tuning on individual corpora improves the scores (Hotel seems to be the most challenging corpus); a model trained solely on PDTB obtains the lowest scores across all corpora, while Discovery and AD perform on par in PEC and MTX, but the model trained on AD stands out with higher scores on Hotel. 
The scores obtained after fine-tuning on individual corpora are superior to the reference values reported for the original data (except for PDTB on PEC), meaning that the Seq2Seq models successfully increase the coverage of ADUs being preceded by explicit DMs (even departing from the input data deprived of DMs, i.e., ``Input data: removed DMs'' setting). 
%
%
Combining the corpora positively impacts the scores on Hotel (23 percentage points above best individual results), with similar scores obtained on PEC and MTX. 
Surprisingly, \Tlarge{} again obtains lower scores. 

%
%
For the ``Input data: original'' setting, we again obtain higher scores. 
These improvements are smaller for Hotel because the original text is mostly deprived of explicit DMs. 
Finally, we observe that in this setting, we can obtain very high coverage scores across all corpora: $98\%$ for PEC, $95\%$ for MTX, and $69\%$ for Hotel.

\subsection{Comparison with ChatGPT}

ChatGPT\footnote{\url{https://openai.com/blog/chatgpt/}} \citep{Leiter2023ChatGPTAM} is a popular large language model built on top of the GPT-3.5 series~\citep{BrMaRy20} and optimized to interact in a conversational way.  
Even though ChatGPT is publicly available, interacting with it can only be done with limited access. Consequently, we are unable to conduct large-scale experiments and fine-tune the model on specific tasks. 
To compare the zero-shot performance of ChatGPT in our task, we run a small-scale experiment and compare the results obtained with the models presented in Section~\ref{ssec:e2e_dm_augment_results}. 
We sampled 11 essays from the test set of the PEC (totaling 60 paragraphs) for this small-scale experiment, following the same evaluation protocol described in Section~\ref{ssec:e2e_dm_augment_eval_protocol}. 
Detailed results can be found in Appendix~\ref{app:e2e_dm_augment_results_chatgpt}. 
Even though our fine-tuned models surpass ChatGPT in most of the metrics (except for ``disc rel''), especially in terms of coverage, it is remarkable that ChatGPT, operating in a zero-shot setting, is competitive. 
With some fine-tuning, better prompt-tuning or in-context learning, we believe that ChatGPT (and similar LLMs) might excel in the proposed DM augmentation task.

%% file: structure/06_downstream.tex
\section{Downstream task evaluation}
\label{sec:downstream_task}

We assess the impact of the end-to-end DM augmentation approach detailed in Section~\ref{sec:e2e_dm_augment} on an \ArgMin{} downstream task, namely: ADU identification and classification. 
We operate on the token level with the label set: 
$\{O\} \cup \{B, I\} \times T$, where $T$ is a corpus-specific set of ADU labels. 
For instance, for PEC, $T = \{\textit{\Premise{}}, \textit{\Claim{}}, \textit{\MajorClaim{}}\}$. 
This subtask is one of the most fundamental in the \ArgMin{} process and considered in many other \ArgMin{} studies~\citep{eger-etal-2018-cross,schulz-etal-2018-multi,MoOzMo22}. 
Its reduced complexity compared to tasks that also include relation identification makes a subsequent analysis of the impact of the proposed approach easier.  

\subsection{Experimental setup}

\begin{figure}[t]
	\centerline{\includegraphics[width=0.55\textwidth,keepaspectratio]{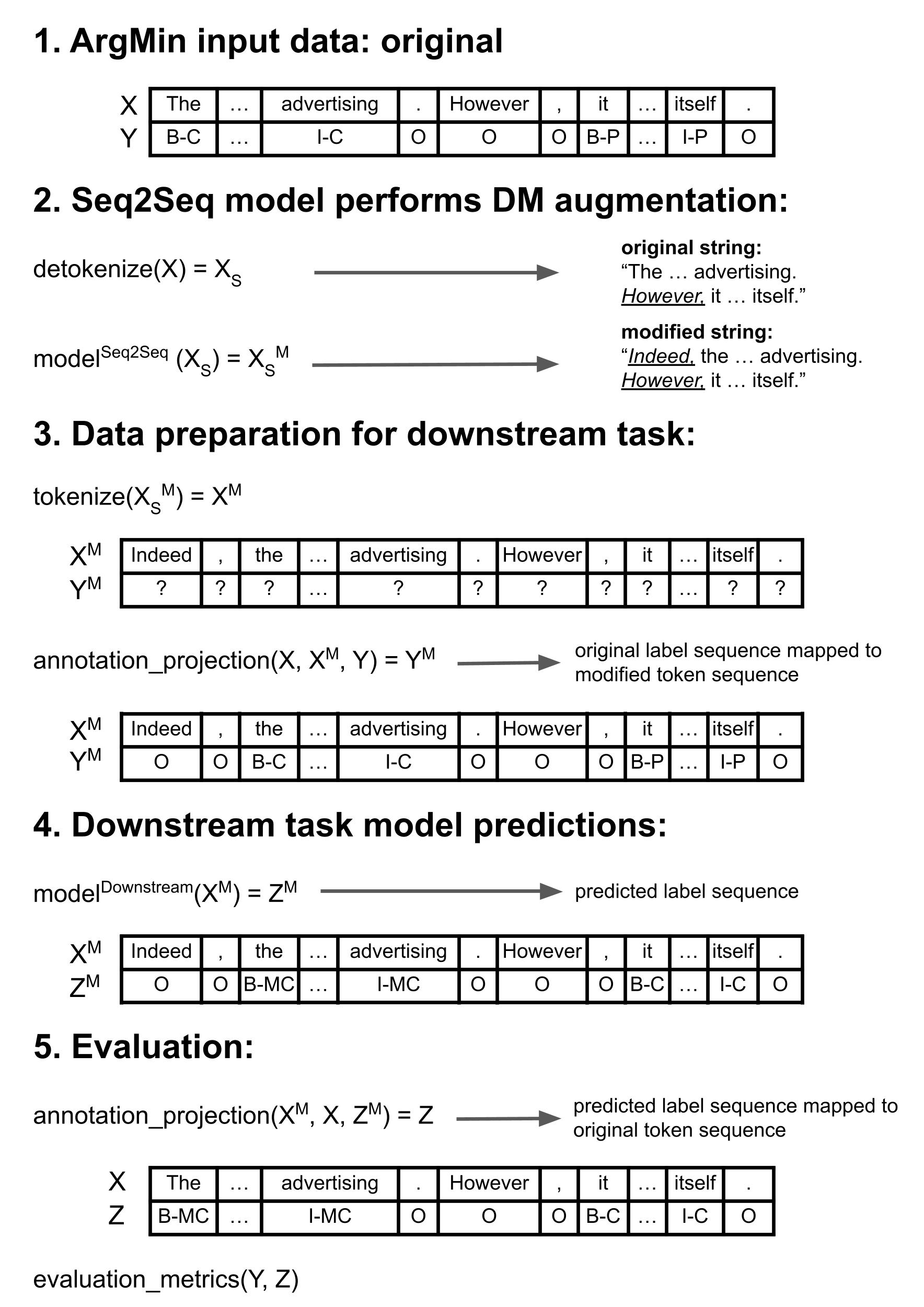}}
	\caption{Downstream task experimental setup process.}
	\label{fig:downstream_task_exp_setup_process}
\end{figure}

We assess the impact of the proposed DM augmentation approach in the downstream task when the Seq2Seq models (described in Section~\ref{sec:e2e_dm_augment}) are asked to augment a text based on two versions (similar to Section~\ref{ssec:e2e_dm_augment_results}) of the input data: 
(a) the original input sequence (``Input data: original''); 
(b) the input sequence obtained after the removal of all explicit DMs (``Input data: removed DMs'').  

In Figure~\ref{fig:downstream_task_exp_setup_process}, we illustrate the experimental setup process using ``Input data: original'' (step 1), where $X$ corresponds to the original token sequence and $Y$ to the original label sequence (as provided in the gold \ArgMin{} annotations). 
The process is similar for ``Input data: removed DMs''. 

In step 2 (Fig.~\ref{fig:downstream_task_exp_setup_process}), a Seq2Seq model performs DM augmentation. 
Since Seq2Seq models work with strings and the input data is provided as token sequences, we need to detokenize the original token sequence (resulting in $X_{S}$ in Fig.~\ref{fig:downstream_task_exp_setup_process}). All tokenization and detokenization operations are performed using sacremoses. 
At the end of this step, we obtain the output provided by the Seq2Seq model (i.e., $X_{S}^{M}$, the text augmented with DMs), which will be used as the input data for the downstream task model (the model trained and evaluated on the downstream task) in the following steps. 

Given that the output provided by the Seq2Seq model is different from the original token sequence $X$ (based on which we have the gold \ArgMin{} annotations), we need to map the original label sequence ($Y$) to the modified token sequence (i.e., $X^{M}$, the token sequence obtained after tokenization of the Seq2Seq output string $X_{S}^{M}$). To this end, in step 3 (Fig.~\ref{fig:downstream_task_exp_setup_process}), we employ an annotation projection procedure, detailed in Appendix~\ref{app:ann_proj}. 
Based on this annotation projection procedure, we train the downstream model using the modified token sequence ($X^{M}$) and the corresponding label sequence obtained via annotation projection (i.e., $Y^{M}$, the original label sequence mapped to the modified token sequence). 
Then, using the trained model, we obtain, in step 4 (Fig.~\ref{fig:downstream_task_exp_setup_process}), the predictions for the test set (i.e., $Z^{M}$, which also contains modified sequences). 

For a fair comparison between different approaches, in step 5 (Fig.~\ref{fig:downstream_task_exp_setup_process}), we map back the predicted label sequence ($Z^{M}$) to the original token sequence (i.e., $Z$ corresponds to the predicted label sequence mapped to the original token sequence), using the same annotation projection procedure.  
Consequently, despite all the changes made by the Seq2Seq model, we ensure that the downstream task evaluation is performed on the same grounds for each approach. 
This is crucial to obtain insightful token-level and component-level (i.e., ADUs in this downstream task) metrics. 
As evaluation metrics, we use the following:
(a) seqeval\footnote{\url{https://huggingface.co/spaces/evaluate-metric/seqeval}} is a popular framework for sequence labeling evaluation typically used to evaluate the performance on chunking tasks such as named-entity recognition and semantic role labeling; 
(b) flat token-level macro-F1 as implemented in scikit-learn~\citep{PeVaGr11}. 

For the downstream task, we employ a BERT model (``bert-base-cased'') following a typical sequence labeling approach. 
We use default ``AutoModelForTokenClassification''  and ``TrainingArguments'' parameters provided by the HuggingFace library, except for the following: learning rate = $2e-5$, weight decay = $0.01$, max training epochs = $50$, and evaluation metric (to select the best epoch based on the dev set) is token-level macro f1-score (similar to prior work, e.g., \citep{schulz-etal-2018-multi}).

\subsection{Results}
\label{ssec:downstream_task_results}

Table~\ref{tab:adu_bd_and_class_results} shows the results obtained for the downstream task. 
The line identified with a ``none'' in the column ``DM augmentation model'' refers to the scores obtained by a baseline model in the corresponding input data setting, in which the downstream model receives as input the data without the intervention of any DM augmentation Seq2Seq model; for the remaining lines, the input sequence provided to the downstream model was previously augmented using a pre-trained Seq2Seq model (\BARTbase{}, \Tbase{}, or \Tlarge{}) fine-tuned on the combination of the three corpora described in Section~\ref{ssec:data_dm_annotations} (``Discovery + AD + PDTB''). 
For ``Input data: original'', the scores in ``none'' correspond to the current state-of-the-art using a BERT model. 
For ``Input data: removed DMs'', the scores in ``none'' correspond to a very challenging setup for sequence labeling models because they are asked to perform the downstream task without explicit signals. 

\begin{table}[t]
\centering
\footnotesize
\begin{tabular}{ccccccccccc}
\hline
\multicolumn{2}{c|}{} & \multicolumn{3}{c|}{PEC} & \multicolumn{3}{c|}{MTX} & \multicolumn{3}{c}{Hotel} \\ \hline
\multicolumn{2}{c|}{\begin{tabular}[c]{@{}c@{}}DM augmentation\\ model\end{tabular}} & \begin{tabular}[c]{@{}c@{}}seqeval\\ f1\end{tabular} & \begin{tabular}[c]{@{}c@{}}token\\ acc\end{tabular} & \multicolumn{1}{c|}{\begin{tabular}[c]{@{}c@{}}token\\ macro f1\end{tabular}} & \begin{tabular}[c]{@{}c@{}}seqeval\\ f1\end{tabular} & \begin{tabular}[c]{@{}c@{}}token\\ acc\end{tabular} & \multicolumn{1}{c|}{\begin{tabular}[c]{@{}c@{}}token\\ macro f1\end{tabular}} & \begin{tabular}[c]{@{}c@{}}seqeval\\ f1\end{tabular} & \begin{tabular}[c]{@{}c@{}}token\\ acc\end{tabular} & \begin{tabular}[c]{@{}c@{}}token\\ macro f1\end{tabular} \\ \hline \hline
\multicolumn{11}{l}{\textit{Input data: removed DMs}} \\ \hline
\multirow{2}{*}{(1)} & \multicolumn{1}{c|}{\multirow{2}{*}{none}} & .6241 & \textbf{.8294} & \multicolumn{1}{c|}{.7457} & .6922 & .8993 & \multicolumn{1}{c|}{.8409} & \textbf{.3341} & .5016 & \textbf{.4855} \\
 & \multicolumn{1}{c|}{} & \scriptsize ($\pm$ .007) & \scriptsize ($\pm$ .005) & \multicolumn{1}{c|}{\scriptsize ($\pm$ .003)} & \scriptsize ($\pm$ .029) & \scriptsize ($\pm$ .005) & \multicolumn{1}{c|}{\scriptsize ($\pm$ .011)} & \scriptsize ($\pm$ .013) & \scriptsize ($\pm$ .018) & \scriptsize ($\pm$ .003) \\
\multirow{2}{*}{(2)} & \multicolumn{1}{c|}{\multirow{2}{*}{BART-base}} & .6301 & .8252 & \multicolumn{1}{c|}{.7435} & .7114 & .8995 & \multicolumn{1}{c|}{.8437} & .2703 & .4907 & .4637 \\
 & \multicolumn{1}{c|}{} & \scriptsize ($\pm$ .009) & \scriptsize ($\pm$ .004) & \multicolumn{1}{c|}{\scriptsize ($\pm$ .007)} & \scriptsize ($\pm$ .025) & \scriptsize ($\pm$ .015) & \multicolumn{1}{c|}{\scriptsize ($\pm$ .018)} & \scriptsize ($\pm$ .008) & \scriptsize ($\pm$ .004) & \scriptsize ($\pm$ .009) \\
\multirow{2}{*}{(3)} & \multicolumn{1}{c|}{\multirow{2}{*}{T5-base}} & \textbf{.6397} & .8247 & \multicolumn{1}{c|}{.7438} & \textbf{.7491} & .9129 & \multicolumn{1}{c|}{.8630} & .3121 & \textbf{.5540} & .4764 \\
 & \multicolumn{1}{c|}{} & \scriptsize ($\pm$ .007) & \scriptsize ($\pm$ .003) & \multicolumn{1}{c|}{\scriptsize ($\pm$ .002)} & \scriptsize ($\pm$ .033) & \scriptsize ($\pm$ .007) & \multicolumn{1}{c|}{\scriptsize ($\pm$ .008)} & \scriptsize ($\pm$ .022) & \scriptsize ($\pm$ .028) & \scriptsize ($\pm$ .019) \\
\multirow{2}{*}{(4)} & \multicolumn{1}{c|}{\multirow{2}{*}{T5-large}} & .6284 & .8258 & \multicolumn{1}{c|}{\textbf{.7459}} & .6988 & \textbf{.9140} & \multicolumn{1}{c|}{\textbf{.8671}} & .2902 & .5038 & .4307 \\
 & \multicolumn{1}{c|}{} & \scriptsize ($\pm$ .020) & \scriptsize ($\pm$ .011) & \multicolumn{1}{c|}{\scriptsize ($\pm$ .010)} & \scriptsize ($\pm$ .067) & \scriptsize ($\pm$ .017) & \multicolumn{1}{c|}{\scriptsize ($\pm$ .021)} & \scriptsize ($\pm$ .020) & \scriptsize ($\pm$ .019) & \scriptsize ($\pm$ .013) \\ \hline \hline
\multicolumn{11}{l}{\textit{Input data: original}} \\ \hline
\multirow{2}{*}{(A)} & \multicolumn{1}{c|}{\multirow{2}{*}{none}} & \textbf{.6785} & \textbf{.8557} & \multicolumn{1}{c|}{\textbf{.7871}} & .7489 & \textbf{.9193} & \multicolumn{1}{c|}{.8674} & .3120 & .4928 & \textbf{.4948} \\
 & \multicolumn{1}{c|}{} & \scriptsize ($\pm$ .013) & \scriptsize ($\pm$ .003) & \multicolumn{1}{c|}{\scriptsize ($\pm$ .004)} & \scriptsize ($\pm$ .037) & \scriptsize ($\pm$ .009) & \multicolumn{1}{c|}{\scriptsize ($\pm$ .019)} & \scriptsize ($\pm$ .010) & \scriptsize ($\pm$ .009) & \scriptsize ($\pm$ .023) \\
\multirow{2}{*}{(B)} & \multicolumn{1}{c|}{\multirow{2}{*}{BART-base}} & .6569 & .8466 & \multicolumn{1}{c|}{.7758} & .7398 & .9150 & \multicolumn{1}{c|}{\textbf{.8676}} & \textbf{.3345} & \textbf{.5021} & .4762 \\
 & \multicolumn{1}{c|}{} & \scriptsize ($\pm$ .022) & \scriptsize ($\pm$ .006) & \multicolumn{1}{c|}{\scriptsize ($\pm$ .004)} & \scriptsize ($\pm$ .021) & \scriptsize ($\pm$ .004) & \multicolumn{1}{c|}{\scriptsize ($\pm$ .009)} & \scriptsize ($\pm$ .008) & \scriptsize ($\pm$ .008) & \scriptsize ($\pm$ .013) \\
\multirow{2}{*}{(C)} & \multicolumn{1}{c|}{\multirow{2}{*}{T5-base}} & .6486 & .8396 & \multicolumn{1}{c|}{.7699} & .7550 & .9136 & \multicolumn{1}{c|}{.8554} & .3183 & .4841 & .4666 \\
 & \multicolumn{1}{c|}{} & \scriptsize ($\pm$ .014) & \scriptsize ($\pm$ .010) & \multicolumn{1}{c|}{\scriptsize ($\pm$ .010)} & \scriptsize ($\pm$ .014) & \scriptsize ($\pm$ .012) & \multicolumn{1}{c|}{\scriptsize ($\pm$ .004)} & \scriptsize ($\pm$ .043) & \scriptsize ($\pm$ .004) & \scriptsize ($\pm$ .023) \\
\multirow{2}{*}{(D)} & \multicolumn{1}{c|}{\multirow{2}{*}{T5-large}} & .6264 & .8294 & \multicolumn{1}{c|}{.7556} & \textbf{.7719} & .9157 & \multicolumn{1}{c|}{.8665} & .2885 & .4645 & .4443 \\
 & \multicolumn{1}{c|}{} & \scriptsize ($\pm$ .020) & \scriptsize ($\pm$ .009) & \multicolumn{1}{c|}{\scriptsize ($\pm$ .007)} & \scriptsize ($\pm$ .021) & \scriptsize ($\pm$ .001) & \multicolumn{1}{c|}{\scriptsize ($\pm$ .006)} & \scriptsize ($\pm$ .008) & \scriptsize ($\pm$ .027) & \scriptsize ($\pm$ .011) \\ \hline
\end{tabular}
\caption{ADU boundary detection and classification results}
\label{tab:adu_bd_and_class_results}
\end{table}

We start our analysis by comparing the baseline models, i.e., rows $(1)$ and $(A)$. 
When the DMs are removed from the data, row $(1)$, we observe an expected drop in the scores in all the metrics on the PEC and MTX corpora: the task is more challenging without DMs. 
Given that the original data in the Hotel corpus is already scarce regarding DMs, we observe slightly lower scores for the metric ``token macro f1'' with the removal of DMs. Surprisingly, we observe higher scores (by a small margin) for the remaining metrics. 

Most of the results obtained from DM augmentation models that receive as input the original data (``Input data: original''; rows $(A, B, C, D)$) are superior to the scores obtained in the setting ``Input data: removed DMs'' (rows $(1, 2, 3, 4)$). 
However, even though in Section~\ref{sec:e2e_dm_augment} we observed clear improvements in all metrics for the setting ``Input data: original'', these improvements are reflected with limited effect in the downstream task. 

Comparing row $(1)$, which is the baseline with DMs removed, to rows $(2, 3, 4)$, which give the results after adding DMs with the DM augmentation models previously described, we observe: 
(i) consistent improvements for the MTX dataset, i.e., the results of $(2, 3, 4)$ are better than $(1)$ in all cases; 
(ii) for PEC, all rows $(1, 2, 3, 4)$ have mostly similar scores across the metrics; 
(iii) only \Tbase{} clearly improves, and only for ``token accuracy'', over $(1)$ for Hotel. 

Comparing row $(A)$, which is the baseline with original data, to $(B, C, D)$, which give the results after performing DM augmentation on top of the original data with the Seq2Seq models previously described, we observe: 
(i) the most consistent improvements are obtained again for the MTX dataset, where we observe a 3 percentage point improvement in ``seqeval f1'' for \Tlarge{} over the baseline; 
(ii) \BARTbase{} improves upon the baseline for Hotel according to 2 out of 3 metrics; 
(iii) there are no improvements for PEC, the baseline performs better according to all three metrics. 

To summarize, we observe that in a downstream task, augmenting DMs automatically with recent LMs can be beneficial in some, but not all, cases. 
We believe that with the advent of larger LMs (such as the recent ChatGPT), the capability of these models to perform DM augmentation can be decisively improved in the near future, with a potential impact on downstream tasks (such as \ArgMin{} tasks). 
Overall, our findings not only show the impact of a DM augmentation approach for a downstream \ArgMin{} task but also demonstrate how the proposed approach can be employed to improve the readability and transparency of argument exposition (i.e., introducing explicit signals that clearly unveil the presence and positioning of the ADUs conveyed in the text). 

Finally, we would like to reinforce that the DM augmentation models were fine-tuned on datasets (i.e., ``Discovery + AD + PDTB'') that 
(a) were not manually annotated for \ArgMin{} tasks and 
(b) are different from the downstream task evaluation datasets (i.e., PEC, MTX, and Hotel). 
Consequently, our results indicate that DM augmentation models can be trained on external data to automatically add useful DMs (in some cases, as previously detailed) for downstream task models, despite the differences in the DM augmentation training data and the downstream evaluation data (e.g., domain shift).

\subsection{Error analysis}  

We manually sampled some data instances from each \ArgMin{} corpora (Section~\ref{ssec:data_argmin}) and analyzed the predictions (token-level, for the ADU identification and classification task) made by the downstream task models. 
Furthermore, we also analyzed the DMs automatically added by the Seq2Seq models, assessing whether it is possible to find associations between the DMs that precede the ADUs and the corresponding ADU labels. 

In Appendix~\ref{app:downstream_task_error_analysis}, we provide a detailed analysis for each corpus and show some examples. 
Overall, we observe that DM augmentation models performed well in terms of coverage, augmenting the text with DMs at appropriate locations (i.e., preceding the DMs). This observation is in line with the conclusions taken from the ``Coverage analysis'' in Section~\ref{ssec:e2e_dm_augment_results}. 
However, we observed that the presence of some DMs that are commonly associated as indicators of specific ADU labels (e.g., ``because'' and ``moreover'' typically associated to \Premise{}) are not consistently used by the downstream model to predict the corresponding ADU label accordingly (i.e., the predicted ADU label varies in the presence of these DMs). 
We attribute this to the lack of consistency (we observed, for all corpora, that some DMs are associated to different ADU labels) and variability (e.g.,\ on PEC, in the presence of augmented DMs, the label \MajorClaim{} does not contain clear indicators; in the original text, these indicators are available and explored by the downstream model) of augmented DMs. 
We conclude that these limitations in the quality of the predictions provided by the DM augmentation models conditioned the association of DMs and ADU labels that we expected to be performed by the downstream model. 

Based on this analysis, our assessment is that erroneous predictions of DMs might interfere with the interpretation of the arguments exposed in the text (and, in some cases, might even mislead the downstream model). 
This is an expected drawback from a pipeline architecture (i.e., end-to-end DM augmentation followed by the downstream task). 
However, on the other hand, the potential of DM augmentation approaches is evident, as the presence of coherent and grammatically correct DMs can clearly improve the readability of the text and of argument exposition in particular (as illustrated in the detailed analysis provided in Appendix~\ref{app:downstream_task_error_analysis}).

%% file: structure/02_related_work.tex
\section{Related Work}

\paragraph{Argument mining}

Given the complexity of the task, it is common to divide the \ArgMin{} task in a set of subtasks~\cite{peldszus2015annotated}, namely: ADU identification, ADU classification (e.g.,  premise vs.\ claim), Argumentative Relation Identification (ARI, e.g., link vs.\ no-link), and Argumentative Relation Classification (ARC, e.g., support vs.\ attack). 
In this paper, we focus on ADU identification and classification as downstream tasks (Section~\ref{sec:downstream_task}). 
The standard BiLSTM with a CRF output layer emerged as the state-of-the-art architecture for token-level sequence tagging, including argument mining~\citep{eger-etal-2017-neural,schulz-etal-2018-multi,chernodub-etal-2019-targer}. 
Current state-of-the-art on ADU identification and classification employs BERT~\citep{devlin-etal-2019-bert} or Longformer~\citep{BePeCo20} as base encoders (in some cases, with a CRF layer on top), typically accompanied with specific architectures to tackle a target corpus or task-specific challenges~\citep{wang-etal-2020-argumentation,ye-teufel-2021-end,MoOzMo22,dutta-etal-2022-unsupervised,HaFaRe22}. 
We follow these recent trends by employing a BERT-based sequence labeling model. Since our goal is to assess the impact of the proposed DM augmentation approach, we keep the architecture as simple and generic as possible (standard BERT encoder with a token classification head), but competitive with recent state-of-the-art (as detailed in Section~\ref{sec:downstream_task}). 

Some prior work also studies \ArgMin{} across different corpora.  
Given the variability of annotation schemas, dealing with different conceptualizations (such as tree vs.\ graph-based structures, ADU and relation labels, ADU boundaries, among others) is a common challenge~\citep{CoCaVi20,GaLiTo21,bao-etal-2021-neural}. 
Besides the variability of annotated resources, \ArgMin{} corpora tend to be small~\citep{MoOzMo22}.
To overcome these challenges, some approaches explored transfer learning: 
(a) across different \ArgMin{} corpora~\citep{schulz-etal-2018-multi,putra-etal-2021-multi,MoOzMo22}; 
(b) from auxiliary tasks, such as discourse parsing~\citep{accuosto-saggion-2019-transferring} and fine-tuning pre-trained LMs on large amounts of unlabeled discussion threads from Reddit~\citep{dutta-etal-2022-unsupervised}; and 
(c) from \ArgMin{} corpora in different languages \citep{eger-etal-2018-cross,toledo-ronen-etal-2020-multilingual,eger-etal-2018-pd3,rocha-etal-2018-cross}. 
Exploring additional training data is pointed out as beneficial across different subtasks, especially under low-resource settings; however, domain-shift and differences in annotation schemas are typically referred to as the main challenges. 
Our approach differs by proposing DM augmentation to improve the ability of \ArgMin{} models across different genres, without requiring to devise transfer learning approaches to deal with different annotation schemas: given that the DM augmentation follows a text-to-text approach, we can employ corpus-specific models to address the task for each corpus.

\paragraph{The role of discourse context}

As a discourse parsing task, prior work on \ArgMin{} looked at the intersection between argumentation structures and existing discourse parsing theories (e.g., RST, PDTB), with several studies pointing out that improvements can be obtained for \ArgMin{} tasks by incorporating insights from related discourse parsing tasks~\citep{hewett-etal-2019-utility,huber-etal-2019-aligning}. 
From the state-of-the-art in discourse parsing tasks, it is well known that discourse markers play an important role as strong indicators for discourse relations~\cite{Marcu00,braud-denis-2016-learning}. 
In the field of \ArgMin{}, such lexical clues have also been explored in prior work, either via handcrafted features~\cite{stab-gurevych-2017-parsing,opitz-frank-2019-dissecting} or encoding these representations in neural-based architectures~\cite{kuribayashi-etal-2019-empirical,bao-etal-2021-neural,RoCa22}. 
Including DM in their span representations, \citet{bao-etal-2021-neural} report state-of-the-art results for ADU classification, ARI, and ARC.  
These works rely on the presence of explicit DMs anteceding ADUs, which is a viable assumption for some of the \ArgMin{} corpora containing texts written in English. 
To obtain a system that is robust either in the presence or absence of such lexical clues, we propose to automatically augment the text with the missing DMs using state-of-the-art Seq2Seq models. 
Our proposal complements prior work findings (e.g., including DMs in span representations improves performance across different subtasks) as we propose a text-to-text approach that can be employed to augment the input text provided to state-of-the-art \ArgMin{} models.  

Aligned with our proposal, \citet{Opitz19} frames ARC as a plausibility ranking prediction task. The notion of plausibility comes from adding DMs (from a handcrafted set of 4 possible DM pairs) of different categories (support and attack) between two ADUs and determining which of them is more plausible.  
They report promising results for this subtask, demonstrating that explicitation of DMs can be a feasible approach to tackle some \ArgMin{} subtasks. 
We aim to go one step further by: (a) employing language models to predict plausible DMs (instead of using a handcrafted set of DMs) and  (b) proposing a more realistic DM augmentation scenario, where we receive as input raw text and we do not assume that the ADU boundaries are known. 

However, relying on these DMs also has downsides. 
In a different line of work, \citet{opitz-frank-2019-dissecting} show that the models they employ to address the task of ARC tend to focus on DMs instead of the actual ADU content. They argue that such a system can be easily fooled in cross-document settings (i.e., ADUs belonging to a given argument can be retrieved from different documents), proposing a context-agnostic model that is constrained to encode only the actual ADU content as an alternative. 
We believe that our approach addresses these concerns as follows: 
(a) for the \ArgMin{} tasks addressed in this work, arguments are constrained to document boundaries (cross-document settings are out of scope); 
(b) given that the DM augmentation models are automatically employed for each document, we hypothesize that the models will take into account the surrounding context and adapt the DMs predictions accordingly (consequently, the downstream model can rely on them).  

\paragraph{Explicit vs.\ Implicit relations in discourse parsing}

In discourse parsing, it is well-known that there exists a clear gap between explicit (relations that are marked explicitly with a DM) and implicit (relation between two spans of text exists, but is not marked explicitly with a DM) relation classification, namely, $90\%$ vs. $50\%$ of accuracy (respectively) in 4-way classification (as indicated by \citet{shi-demberg-2019-learning}).  
To improve discourse relation parsing, several works focused on enhancing their systems for implicit relation classification: 
removing DMs from explicit relations for implicit relation classification data augmentation~\citep{braud-denis-2014-combining,rutherford-xue-2015-improving}; 
framing explicit vs.\ implicit relation classification as a domain adaptation problem~\citep{qin-etal-2017-adversarial,huang-li-2019-unsupervised}; 
learning sentence representations by exploring automatically collected large-scale datasets~\citep{nie-etal-2019-dissent,sileo-etal-2019-mining}; 
multi-task learning~\citep{lan-etal-2017-multi,nguyen-etal-2019-employing}; 
automatic explicitation of implicit DMs followed by explicit relation classification~\citep{shi-demberg-2019-learning,kurfali-ostling-2021-lets}. 

To close the gap between explicit and implicit DMs, our approach follows the line of work on explicitation. 
However, we work in a more challenging scenario, where the DM augmentation and downstream tasks are performed at the paragraph level (i.e., from raw text instead of a sentence-pair classification task that assumes that the ADUs are given). 

%% file: structure/07_conclusion.tex
\section{Conclusions}

In this paper, we propose to automatically augment a text with DMs to improve the robustness of \ArgMin{} systems across different genres. 

First, we describe a synthetic template-based test suite created to assess the capabilities of recent LMs to predict DMs and whether LMs are sensitive to specific semantically-critical edits in the text. 
We show that LMs underperform this task in a zero-shot setting, but the performance can be improved after some fine-tuning. 

Then, we assess whether LMs can be employed to automatically augment a text with coherent and grammatically correct DMs in an end-to-end setting. 
We collect a heterogeneous collection of DM-related datasets and show that fine-tuning LMs in this collection improves the ability of LMs in this task.  

Finally, we evaluate the impact of augmented DMs performed by the proposed end-to-end DM augmentation models on the performance of a downstream model (across different \ArgMin{} corpora). 
We obtained mixed results across different corpora. 
Our analysis indicates that DM augmentation models performed well in terms of coverage; however, the lack of consistency and variability of the augmented DMs conditioned the association of DMs and ADU labels that we expected to be performed by the downstream model. 

In future work, we would like to assess how recent LLMs perform in these tasks. Additionally, we would like to increase and improve the variability and quality of the heterogeneous collection of data instances used to fine-tune the end-to-end DM augmentation models (possibly including data related to \ArgMin{} tasks that might inform the models about DMs that are more predominant in \ArgMin{} domains), as improving in this axis might have a direct impact in the downstream task performance. 

We believe that our findings are evidence of the potential of DM augmentation approaches. DM augmentation models can be deployed to improve the readability and transparency of arguments exposed in written text, such as embedding this approach in assistive writing tools. 

%% file: structure/99_limitations.tex
\section{Limitations}

One of the anchors of this work is evidence from prior work that DMs can play an important role to identify and classify ADUs; prior work is mostly based on DMs preceding the ADUs. 
Consequently, we focus on DMs preceding the ADUs. 
We note that DMs following ADUs might also occur in natural language and might be indicative of ADU roles. 
However, this phenomenon is less frequent in natural language and also less studied in related work \citep{eger-etal-2018-cross,kuribayashi-etal-2019-empirical,sileo-etal-2019-mining}. 

The Artificial Dataset proposed in Section~\ref{ssec:artificial_dataset} follows a template-based approach, instantiated with examples extracted from the CoPAs provided by \citet{bilu-etal-2019-argument}. 
While some control over linguistic phenomena occurring in the dataset was important to investigate our hypothesis, the downside is a lack of diversity. 
Nonetheless, we believe that the dataset contains enough diversity for the purposes studied in this work (e.g., different topics, several parameters that result in different sentence structures, etc.). Future work might include expanding the dataset with more templates and data instances. 

Our proposed approach follows a pipeline architecture: end-to-end DM augmentation followed by the downstream task. Consequently, erroneous predictions made by the DM augmentation model might mislead the downstream task model. 
Furthermore, the end-to-end DM augmentation employs a Seq2Seq model. 
Even though these models were trained to add DMs without changing further content, it might happen in some cases that the original ADU content is changed by the model. 
We foresee that, in extreme cases, these edits might lead to a different argument content being expressed (e.g., changing the stance, adding/removing negation expressions, etc.); however, we note that we did not observe this in our experiments. 
In a few cases, we observed minor edits being performed to the content of the ADUs, mostly related to grammatical corrections. 

We point out that despite the limited effectiveness of the proposed DM augmentation approach in improving the downstream task scores in some settings, our proposal is grounded on a well-motivated and promising research hypothesis, solid experimental setup, and detailed error analysis that we hope can guide future research. 
Similar to recent trends in the community (Insights NLP workshop~\citep{insights-2022-insights}, ICBINB Neurips workshop and initiative\footnote{\url{http://icbinb.cc/}}, etc.), we believe that well-motivated and well-executed research can also contribute to the progress of science, going beyond the current emphasis on state-of-the-art results.

%% file: structure/appendices.tex
\section{Artificial dataset - templates}
\label{app:art_dataset_templates}

The templates are based on a set of configurable parameters, namely: 
\begin{itemize}
    \item number of ADUs $\in \{2, 3\}$: sample might contain 2 ADUs following the structure ``$\textit{dm}_1$ $X_1$, $\textit{dm}_2$ $X_2$.'', where one of the ADUs ($X_1$ or $X_2$) is a claim and the other a premise; or contain 3 ADUs (claim and both premises) following the structure ``$\textit{dm}_1$ $X_1$, $\textit{dm}_2$ $X_2$. $\textit{dm}_3$, $X_3$.'';
    \item stance role $\in \{original, opposite\}$: each sample contains a single claim, which might employ the ``original'' or ``opposite'' stance;
    \item claim position $\in \{1, 2\}$: the claim is always in the first sentence, either in the beginning ($1$) or end ($2$) of the sentence (to avoid the unusual text sequence where we have two premises in a single sentence followed by an isolated claim in the second sentence);
    \item premise role $\in \{support, attack\}$: only used when ``number of ADUs'' = 2; dictates which of the premises is chosen;
    \item supportive premise position $\in \{1, 2\}$: only used when ``number of ADUs'' = 3, indicates whether the supportive premise should occur before ($1$) the attacking premise or after ($2$); 
    \item prediction type $\in \{\textit{dm}_{1}, \textit{dm}_{2}, \textit{dm}_{3}\}$: let $\textit{dm}_{i}$ be the option chosen, then the mask token will be placed in DM preceding the ADU in position $i$ (if ``number of ADUs'' = 2 only $\textit{dm}_{1}$ and $\textit{dm}_{2}$ are allowed). 
\end{itemize}

\section{Artificial dataset - DMs set}
\label{app:art_dataset_dms}
 
DMs are added based on the role of the ADU that they precede, using the following fixed set of DMs.
If preceding the claim, then we add one the following: ``I think that'', ``in my opinion'', or ``I believe that''. 
For the supportive premise, if in position $\textit{dm}_3$ we add one of the following: ``moreover'', ``furthermore'', or ``indeed''. Otherwise, one of the following: ``because'', ``since'', or ``given that''. 
For the attacking premise, in $\textit{dm}_3$ we add: ``however'', ``on the other hand'', or ``conversely''. Otherwise, ``although'', ``even though'', or ``even if''.

\section{Artificial dataset - instantiation procedure based on CoPAs}
\label{app:art_dataset_copas}

CoPAs are sets of propositions that are often used when debating a recurring theme (e.g., the premises mentioned in Section~\ref{ssec:artificial_dataset} and used in Figure~\ref{fig:artificial_dataset_example} are related to the theme ``Clean energy''). 
For each CoPA, \citet{bilu-etal-2019-argument} provide two propositions that people tend to agree as supporting different points of view for a given theme. 
We use these propositions as supportive and attacking premises towards a given claim. 
Each CoPA is also associated with a set of motions to which the corresponding theme is relevant. 
A motion is defined as a pair $\langle action, \: topic \rangle$, where an action is a term coming from a closed set of allowed actions (e.g., abolish, adopt, legalize, etc.), and a topic is a Wikipedia title. 
For example, for the theme ``Clean energy'', we can find the motion $\langle introduce, \: carbon \: taxes \rangle$, which can be written as ``we should introduce carbon taxes''. 
We use these motions as claims in our Artificial dataset. 
Based on these instantiations and the set of templates, we can generate different samples that resemble real-world arguments. 
The ``opposite stance'' is not provided in the original resources from \citet{bilu-etal-2019-argument}. For a specific motion, we manually selected the action (from the set of allowed actions) that could be employed as ``opposite stance'' (e.g., $\langle abolish, \: carbon \: taxes \rangle$). 

\section{Automatic evaluation metrics}
\label{app:auto_eval_metrics}

The evaluation metrics employed to assess the quality of the predicted DMs are the following: 
\begin{itemize}
    \item word embeddings text similarity (``word embs''): Cosine similarity using an average of word vectors. Based on pre-trained embeddings ``en\_core\_web\_lg'' from Spacy library~\footnote{\url{https://spacy.io/}}; 
    \item retrofitted word embeddings text similarity (``retrofit embs''): Cosine similarity using an average of word vectors. Based on pre-trained embeddings from LEAR~\footnote{\url{https://github.com/nmrksic/lear}}; 
    \item sentence embeddings text similarity (``sbert embs''): we use the pre-trained sentence embeddings ``all-mpnet-base-v2'' from SBERT library~\citep{reimers-gurevych-2019-sentence}, indicated as the model with the highest average performance on encoding sentences over 14 diverse tasks from different domains. To compare gold and predicted DMs representations, we use cosine similarity;
    \item argument marker sense (``arg marker''): list of 115 DMs from \citet{stab-gurevych-2017-parsing}. Senses are divided in the following categories: ``forward'', ``backward'', ``thesis'', and ``rebuttal'' indicators. Each gold and predicted DM is mapped to one of the senses based on a strict lexical match with the list of DMs available for each sense. If the DM is not matched, then we assign the label ``none''. If the gold DM is ``none'', we do not consider this instance in the evaluation (the DM is out of the scope for the list of DMs available in the senses list, so we cannot make a concrete comparison with the predicted DM); 
    \item discourse relation sense (``disc rel''): we use a lexicon of 149 English DMs called ``DiMLex-Eng''~\citep{das-etal-2018-constructing}. These DMs were extracted from PDTB 2.0~\citep{prasad-etal-2008-penn}, RST-SC~\citep{DaTa18}, and Relational Indicator List \citep{BiRa11}. Each DM maps to a set of possible senses. For a given DM, we choose the sense that the DM occurs more frequently. Senses are organized hierarchically in 3 levels (e.g., the DM ``consequently'' is mapped to the sense ``Contingency.Cause.Result''). In this work, we consider only the first level of the senses (i.e., ``Comparison'', ``Contingency'', ``Expansion'', and ``Temporal'') as a simplification and to avoid the propagation of errors between levels (i.e., an error in level 1 entails an error in level 2, and so on). Each prediction is mapped to one of the senses based on a strict lexical match. If the word is not matched, then we assign the label ``none'';
\end{itemize}

\section{Fill-in-the-mask discourse marker prediction - error analysis}
\label{app:fitm_error_analysis_details}

Table~\ref{tab:fitm_edge_cases_gold} shows some instances and the corresponding gold DMs (accompanied by the discourse-level senses ``arg marker'' and ``disc rel'' in parenthesis) from the test set of the Artificial Dataset. 
In this sample, we highlight some of the challenging instances that can be found in the Artificial dataset. 
More concretely, instances with id 1 and 2 belong to the same instantiation of the core elements, but the template used to generate the instances differs in a single parameter (i.e., the parameter ``premise role'' which changes the premise that is presented, requiring the model to predict different DMs); instances 3 and 4 belong to the same instantiation of the core elements, but the template used to generate the instances differs in two parameters (i.e., ``stance role'' that dictates the stance of the claim and ``supportive premise position'' that dictates the position of the supportive premise, requiring the model to predict the same DM in both instances); and so on.  

\begin{table}[h!]
\footnotesize
\centering
\begin{tabular}{clc}
\hline
id & \multicolumn{1}{c}{Masked sentence} & Gold \\ \hline \hline
1 & \begin{tabular}[c]{@{}l@{}}I think that the use of AI should be abandoned, $<$mask$>$ these new technologies \\ are not as reliable as conventional ones.\end{tabular} & \begin{tabular}[c]{@{}c@{}}because\\ \scriptsize \textit{(backward)}\\ \scriptsize \textit{(contingency)}\end{tabular} \\ \hline
2 & \begin{tabular}[c]{@{}l@{}}I think that the use of AI should be abandoned, $<$mask$>$ the use of AI is better than \\ the older options.\end{tabular} & \begin{tabular}[c]{@{}c@{}}although\\ \scriptsize \textit{(rebuttal)}\\ \scriptsize \textit{(comparison)}\end{tabular} \\ \hline
3 & \begin{tabular}[c]{@{}l@{}}I think that the use of AI should be abandoned, although the use of AI is better than \\ the older options. $<$mask$>$, these new technologies are not as reliable\\  as conventional ones.\end{tabular} & \begin{tabular}[c]{@{}c@{}}Moreover\\ \scriptsize \textit{(backward)}\\ \scriptsize \textit{(expansion)}\end{tabular} \\ \hline
4 & \begin{tabular}[c]{@{}l@{}}I think that the use of AI should be encouraged, although these new technologies \\ are not as reliable as conventional ones. $<$mask$>$, the use of AI is better than \\ the older options.\end{tabular} & \begin{tabular}[c]{@{}c@{}}Moreover\\ \scriptsize \textit{(backward)}\\ \scriptsize \textit{(expansion)}\end{tabular} \\ \hline
5 & \begin{tabular}[c]{@{}l@{}}$<$mask$>$ adolescents are as capable as adults, I think that we should increase \\ youth rights. However, many adolescents can not make responsible decisions.\end{tabular} & \begin{tabular}[c]{@{}c@{}}Because\\ \scriptsize \textit{(forward)}\\ \scriptsize \textit{(contingency)}\end{tabular} \\ \hline
6 & \begin{tabular}[c]{@{}l@{}}$<$mask$>$ adolescents are as capable as adults, I think that we should abolish \\ youth rights. Moreover, many adolescents can not make responsible decisions.\end{tabular} & \begin{tabular}[c]{@{}c@{}}Although\\ \scriptsize \textit{(rebuttal)}\\ \scriptsize \textit{(comparison)}\end{tabular} \\ \hline
7 & \begin{tabular}[c]{@{}l@{}}I think that we should increase internet censorship, $<$mask$>$ enforcement \\ tends to be less effective than persuasion and education. Moreover, a decisive \\ and enforced policy is the best way to deliver a message.\end{tabular} & \begin{tabular}[c]{@{}c@{}}although\\ \scriptsize \textit{(rebuttal)}\\ \scriptsize \textit{(comparison)}\end{tabular} \\ \hline
8 & \begin{tabular}[c]{@{}l@{}}I think that we should abandon internet censorship, $<$mask$>$ enforcement \\ tends to be less effective than persuasion and education. However, a decisive \\ and enforced policy is the best way to deliver a message.\end{tabular} & \begin{tabular}[c]{@{}c@{}}because\\ \scriptsize \textit{(backward)}\\ \scriptsize \textit{(contingency)}\end{tabular} \\ \hline
\end{tabular}
\caption{Some instances from the test set of the Artificial Dataset}
\label{tab:fitm_edge_cases_gold}
\end{table}

\begin{table}[h!]
\footnotesize
\centering
\begin{tabular}{c|cccc|c}
\hline
 & \multicolumn{4}{c|}{Zero-shot} & Fine-tuned \\ \hline
id & BERT & XLM-R & BART V1 & BART V2 & BERT (all) \\ \hline \hline
1 & \begin{tabular}[c]{@{}c@{}}as\\ \scriptsize \textit{(none)}\\ \scriptsize \textit{(temporal)}\end{tabular} & \begin{tabular}[c]{@{}c@{}}because\\ \scriptsize \textit{(backward)}\\ \scriptsize \textit{(contingency)}\end{tabular} & \begin{tabular}[c]{@{}c@{}}because\\ \scriptsize \textit{(backward)}\\ \scriptsize \textit{(contingency)}\end{tabular} & \begin{tabular}[c]{@{}c@{}}because\\ \scriptsize \textit{(backward)}\\ \scriptsize \textit{(contingency)}\end{tabular} & \begin{tabular}[c]{@{}c@{}}because\\ \scriptsize \textit{(backward)}\\ \scriptsize \textit{(contingency)}\end{tabular} \\ \hline
2 & \begin{tabular}[c]{@{}c@{}}because\\ \scriptsize \textit{(backward)}\\ \scriptsize \textit{(contingency)}\end{tabular} & \begin{tabular}[c]{@{}c@{}}because\\ \scriptsize \textit{(backward)}\\ \scriptsize \textit{(contingency)}\end{tabular} & \begin{tabular}[c]{@{}c@{}}but\\ \scriptsize \textit{(rebuttal)}\\ \scriptsize \textit{(comparison)}\end{tabular} & \begin{tabular}[c]{@{}c@{}}but I think that \\ it is possible that\\ \scriptsize \textit{(none)}\\ \scriptsize \textit{(none)}\end{tabular} & \begin{tabular}[c]{@{}c@{}}although\\ \scriptsize \textit{(rebuttal)}\\ \scriptsize \textit{(comparison)}\end{tabular} \\ \hline
3 & \begin{tabular}[c]{@{}c@{}}Unfortunately\\ \scriptsize \textit{(none)}\\ \scriptsize \textit{(none)}\end{tabular} & \begin{tabular}[c]{@{}c@{}}However\\ \scriptsize \textit{(rebuttal)}\\ \scriptsize \textit{(comparison)}\end{tabular} & \begin{tabular}[c]{@{}c@{}}However\\ \scriptsize \textit{(rebuttal)}\\ \scriptsize \textit{(comparison)}\end{tabular} & \begin{tabular}[c]{@{}c@{}}However\\ \scriptsize \textit{(rebuttal)}\\ \scriptsize \textit{(comparison)}\end{tabular} & \begin{tabular}[c]{@{}c@{}}Moreover\\ \scriptsize \textit{(backward)}\\ \scriptsize \textit{(expansion)}\end{tabular} \\ \hline
4 & \begin{tabular}[c]{@{}c@{}}However\\ \scriptsize \textit{(rebuttal)}\\ \scriptsize \textit{(comparison)}\end{tabular} & \begin{tabular}[c]{@{}c@{}}However\\ \scriptsize \textit{(rebuttal)}\\ \scriptsize \textit{(comparison)}\end{tabular} & \begin{tabular}[c]{@{}c@{}}However\\ \scriptsize \textit{(rebuttal)}\\ \scriptsize \textit{(comparison)}\end{tabular} & \begin{tabular}[c]{@{}c@{}}However, I think that in the \\ long run, in the short term\\ \scriptsize \textit{(none)}\\ \scriptsize \textit{(none)}\end{tabular} & \begin{tabular}[c]{@{}c@{}}Moreover\\ \scriptsize \textit{(backward)}\\ \scriptsize \textit{(expansion)}\end{tabular} \\ \hline
5 & \begin{tabular}[c]{@{}c@{}}If\\ \scriptsize \textit{(none)}\\ \scriptsize \textit{(contingency)}\end{tabular} & \begin{tabular}[c]{@{}c@{}}Because\\ \scriptsize \textit{(forward)}\\ \scriptsize \textit{(contingency)}\end{tabular} & \begin{tabular}[c]{@{}c@{}}Since\\ \scriptsize \textit{(backward)}\\ \scriptsize \textit{(contingency)}\end{tabular} & \begin{tabular}[c]{@{}c@{}}Since\\ \scriptsize \textit{(backward)}\\ \scriptsize \textit{(contingency)}\end{tabular} & \begin{tabular}[c]{@{}c@{}}Because\\ \scriptsize \textit{(forward)}\\ \scriptsize \textit{(contingency)}\end{tabular} \\ \hline
6 & \begin{tabular}[c]{@{}c@{}}If\\ \scriptsize \textit{(none)}\\ \scriptsize \textit{(contingency)}\end{tabular} & \begin{tabular}[c]{@{}c@{}}Because\\ \scriptsize \textit{(forward)}\\ \scriptsize \textit{(contingency)}\end{tabular} & \begin{tabular}[c]{@{}c@{}}Since\\ \scriptsize \textit{(backward)}\\ \scriptsize \textit{(contingency)}\end{tabular} & \begin{tabular}[c]{@{}c@{}}Since\\ \scriptsize \textit{(backward)}\\ \scriptsize \textit{(contingency)}\end{tabular} & \begin{tabular}[c]{@{}c@{}}Although\\ \scriptsize \textit{(rebuttal)}\\ \scriptsize \textit{(comparison)}\end{tabular} \\ \hline
7 & \begin{tabular}[c]{@{}c@{}}since\\ \scriptsize \textit{(backward)}\\ \scriptsize \textit{(contingency)}\end{tabular} & \begin{tabular}[c]{@{}c@{}}because\\ \scriptsize \textit{(backward)}\\ \scriptsize \textit{(contingency)}\end{tabular} & \begin{tabular}[c]{@{}c@{}}but\\ \scriptsize \textit{(rebuttal)}\\ \scriptsize \textit{(comparison)}\end{tabular} & \begin{tabular}[c]{@{}c@{}}but I think that\\ \scriptsize \textit{(none)}\\ \scriptsize \textit{(none)}\end{tabular} & \begin{tabular}[c]{@{}c@{}}although\\ \scriptsize \textit{(rebuttal)}\\ \scriptsize \textit{(comparison)}\end{tabular} \\ \hline
8 & \begin{tabular}[c]{@{}c@{}}because\\ \scriptsize \textit{(backward)}\\ \scriptsize \textit{(contingency)}\end{tabular} & \begin{tabular}[c]{@{}c@{}}because\\ \scriptsize \textit{(backward)}\\ \scriptsize \textit{(contingency)}\end{tabular} & \begin{tabular}[c]{@{}c@{}}as\\ \scriptsize \textit{(none)}\\ \scriptsize \textit{(temporal)}\end{tabular} & \begin{tabular}[c]{@{}c@{}}as\\ \scriptsize \textit{(none)}\\ \scriptsize \textit{(temporal)}\end{tabular} & \begin{tabular}[c]{@{}c@{}}because\\ \scriptsize \textit{(backward)}\\ \scriptsize \textit{(contingency)}\end{tabular} \\ \hline
\end{tabular}
\caption{Zero-shot and fine-tuned models predictions for some instances sampled from the Artificial Dataset}
\label{tab:fitm_edge_cases_preds}
\end{table}

The predictions made by the zero-shot models and fine-tuned \BERTall{} model (described in Section~\ref{sec:fitm}) for the corresponding instances are shown in Table~\ref{tab:fitm_edge_cases_preds}. 
For example, comparing the predictions made for instances with id 1 and 2, we can observe that both zero-shot \BERT{} and \XLMr{} do not change the semantics of the prediction even though the differences in content require such changes, while both zero-shot BART-based models and \BERTall{} are robust and change the prediction accordingly. 

\section{Human Evaluation - additional details}
\label{app:human_eval_instances}

Table~\ref{tab:human_eval_instances} shows some of the text sequences analyzed in the human evaluation.

\begin{table}[h]
\centering
\footnotesize
\begin{tabular}{lccc}
\hline
\multicolumn{1}{c}{Masked sentence} & Prediction & Gram. & Coh. \\ \hline \hline
\begin{tabular}[c]{@{}l@{}}I think that we should abolish environmental laws, $<$mask$>$ environmentalism \\ stands in the way of technological progress and economic growth.\end{tabular} & because & 1 & 1 \\ \hline
\begin{tabular}[c]{@{}l@{}}I think that we should abolish environmental laws, $<$mask$>$ people must \\ protect nature and respect its biological communities.\end{tabular} & but & 1 & 1 \\ \hline
\begin{tabular}[c]{@{}l@{}}I think that we should abolish environmental laws, $<$mask$>$ people must \\ protect nature and respect its biological communities.\end{tabular} & and that & 1 & 0 \\ \hline
\begin{tabular}[c]{@{}l@{}}I think that we should abolish environmental laws, $<$mask$>$ people must \\ protect nature and respect its biological communities.\end{tabular} & because & 1 & -1 \\ \hline
\begin{tabular}[c]{@{}l@{}}I think that we should introduce goal line technology, $<$mask$>$ it is time \\ to change the old ways and try something new. However, the current system \\ is working, and making such a change could have negative consequences.\end{tabular} & because & 1 & 1 \\ \hline
\begin{tabular}[c]{@{}l@{}}I think that we should introduce goal line technology, $<$mask$>$ the current \\ system is working, and making such a change could have negative consequences.\\ Moreover, it is time to change the old ways and try something new.\end{tabular} & because & 1 & -1 \\ \hline
\begin{tabular}[c]{@{}l@{}}Although it is time to change the old ways and try something new, I think that \\ we should oppose goal line technology. $<$mask$>$, the current system is \\ working, and making such a change could have negative consequences.\end{tabular} & However & 1 & -1 \\ \hline
\begin{tabular}[c]{@{}l@{}}Although it is time to change the old ways and try something new, I think that \\ we should oppose goal line technology. $<$mask$>$, the current system is \\ working, and making such a change could have negative consequences.\end{tabular} & In & -1 & 0 \\ \hline
\begin{tabular}[c]{@{}l@{}}$<$mask$>$ animals should not be treated as property, I think that \\ bullfighting should be legalized.\end{tabular} & While & 1 & 1 \\ \hline
\begin{tabular}[c]{@{}l@{}}$<$mask$>$ animals should not be treated as property, I think that \\ bullfighting should be legalized.\end{tabular} & I think that & 0 & 0 \\ \hline
\begin{tabular}[c]{@{}l@{}}$<$mask$>$ animals should not be treated as property, I think that \\ bullfighting should be legalized.\end{tabular} & Because & 1 & -1 \\ \hline
\end{tabular}
\caption{Some instances used in human evaluation experiments}
\label{tab:human_eval_instances}
\end{table}

\section{End-to-end DM augmentation results - comparison with ChatGPT}
\label{app:e2e_dm_augment_results_chatgpt}

Table~\ref{tab:e2e_dm_augment_results_pec_small_expr} shows the results obtained for the small-scale end-to-end DM augmentation experiment with ChatGPT, including both the explicit DMs accuracy and coverage analysis. 

\begin{table}[h]
\centering
\footnotesize
\begin{tabular}{c|ccccc|c}
\hline
\multicolumn{1}{l|}{} & \multicolumn{5}{c|}{explicit DMs accuracy analysis} & \begin{tabular}[c]{@{}c@{}}coverage\\ analysis\end{tabular} \\ \hline
 & \begin{tabular}[c]{@{}c@{}}word \\ embs\end{tabular} & \begin{tabular}[c]{@{}c@{}}retrofit \\ embs\end{tabular} & \begin{tabular}[c]{@{}c@{}}sbert \\ embs\end{tabular} & \begin{tabular}[c]{@{}c@{}}arg \\ marker\end{tabular} & \begin{tabular}[c]{@{}c@{}}disc \\ rel\end{tabular} & \begin{tabular}[c]{@{}c@{}}pct. DMs\\ predicted\end{tabular} \\ \hline \hline
bart-base (zero-shot) & 0 & 0 & 0 & 0 & 0 & .0163 \\
\multicolumn{1}{l|}{chat-gpt (zero-shot)} & .3630 & .1972 & .2056 & .2222 & \textbf{.3600} & .5476 \\
Discovery & .5269 & .1965 & .2503 & .3819 & .1800 & \textbf{.9443} \\
AD & \textbf{.5807} & \textbf{.2635} & .2697 & .2639 & .2800 & .8626 \\
PDTB & .4009 & .2021 & .1953 & .1875 & .3200 & .6582 \\
\begin{tabular}[c]{@{}c@{}}Discovery + \\ AD + PDTB\end{tabular} & .5400 & .2326 & \textbf{.2937} & \textbf{.4097} & .2400 & .9274 \\ \hline
\end{tabular}
\caption{End-to-end DM augmentation results - small-scale experiment}
\label{tab:e2e_dm_augment_results_pec_small_expr}
\end{table}

\section{Annotation projection}
\label{app:ann_proj}

To map the label sequence from the original sequence to the modified sequence, we implement the Needleman-Wunsch algorithm~\citep{NeWu70}, a well-known sequence alignment algorithm. 
As input, it receives the original and modified token sequences. 
The output is an alignment of the token sequences, token by token, where the goal is to optimize a global score. 
This algorithm might include a special token (the ``gap'' token) in the output sequences.  
Gap tokens are inserted to optimize the alignment of identical tokens in successive sequences. 
The global score attributed to a given alignment is based on a scoring system. We use default values: match score (tokens are identical) = $1$, mismatch score (tokens are different but aligned to optimize alignment sequence) = $-1$, gap penalty (gap token was introduced in one of the sequences) = $-1$. To determine whether two tokens are identical, we use strict lexical match (case insensitive). 
Using the aligned sequences, we map the labels from the original to the modified token sequence. 

\begin{figure}[h]
	\centerline{\includegraphics[width=0.7\textwidth,keepaspectratio]{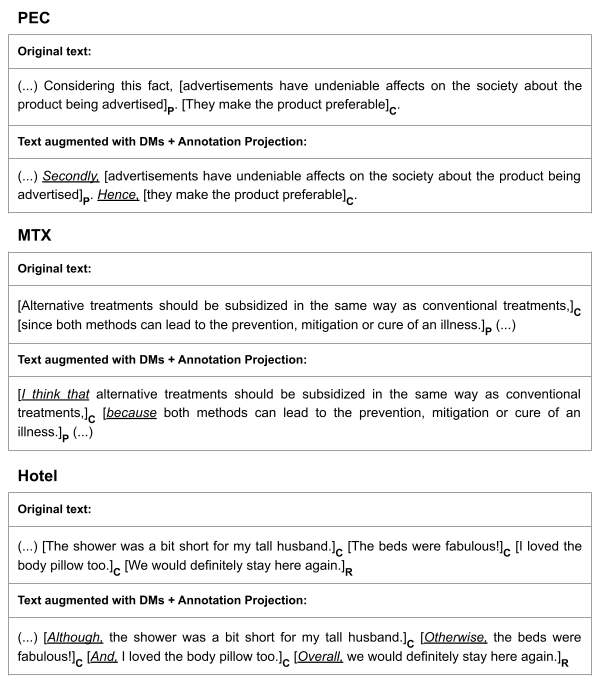}}
	\caption{Annotation projection examples.}
	\label{fig:ann_proj_examples}
\end{figure}

Figure~\ref{fig:ann_proj_examples} illustrates some examples of the output obtained when employing the annotation projection procedure described in this section to the three \ArgMin{} corpora explored in this work. 
``Original text'' corresponds to the original text from which gold annotations for ADU identification and classification are provided in the \ArgMin{} corpora. ``Text augmented with DMs + Annotation Projection'' corresponds to the text obtained after performing DM augmentation (using the pre-trained Seq2Seq model \Tbase{} fine-tuned on the combination of the corpora ``Discovery + AD + PDTB'', as described in Section~\ref{ssec:downstream_task_results}) when we provide as input the text deprived of DMs (i.e., ``Input data: removed DMs'') and the corresponding ADU identification and classification labels obtained after performing annotation projection. 
The underlined text highlights differences between the original and DM augmented text. These differences require a projection of the original label sequence to the label sequence for the corresponding DM augmented text, which is performed using the proposed annotation projection procedure.

\section{Downstream task evaluation - error analysis}
\label{app:downstream_task_error_analysis}

We show some examples of the gold data and predictions made by the downstream task models for the \ArgMin{} corpora explored in this work. 

For each example, we provide: 
\begin{itemize}
    \item ``Gold'': the gold data, including the ADU boundaries (square brackets) and the ADU labels (acronyms in subscript); 
    \item ``Input data: X (Y)'': where ``X'' indicates the version of the input data provided to the DM augmentation model, and ``Y'' indicates whether we perform DM augmentation or not (``none'' indicates that we do not perform DM augmentation and \Tbase{} indicates that we perform DM augmentation using the pre-trained Seq2Seq model \Tbase{} fine-tuned on the combination of the corpora ``Discovery + AD + PDTB''). 
\end{itemize}

\begin{figure}[h]
	\centerline{\includegraphics[width=0.7\textwidth,keepaspectratio]{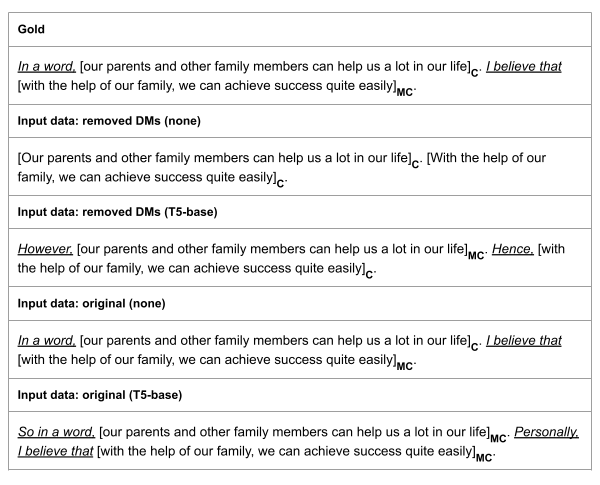}}
	\caption{PEC example containing \Claim{} and \MajorClaim{} in the gold data.}
	\label{fig:pec_e26_p4}
\end{figure}

\begin{figure}[h]
	\centerline{\includegraphics[width=0.7\textwidth,keepaspectratio]{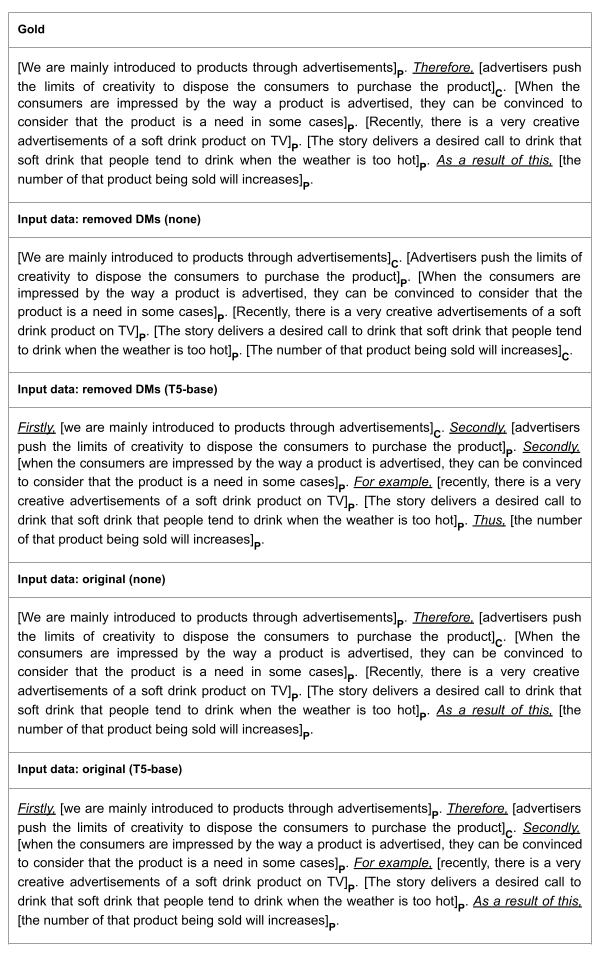}}
	\caption{PEC example containing \Premise{} and \Claim{} in the gold data.}
	\label{fig:pec_e3_p2}
\end{figure}

Figures~\ref{fig:pec_e26_p4} and \ref{fig:pec_e3_p2} show two examples from PEC. 

Figure~\ref{fig:pec_e26_p4} shows a paragraph containing a \Claim{} and \MajorClaim{} in the ``Gold'' data annotations. 
We observe that in the ``Input data: original (none)'' setup, the model predicts \MajorClaim{} frequently in the presence of a DM that can be mapped to the ``arg marker'' sense ``thesis'' (e.g., ``in conclusion'', ``in my opinion'', ``as far as I am concerned'', ``I believe that'', etc.). 
Similar patterns can be observed in the ``Gold'' data annotations. 
We were not able to find similar associations in the ``Input data: removed DMs (\Tbase{})'' setup, for instance. 
As illustrated in ``Input data: removed DMs (none)'', the distinction between \Claim{} and \MajorClaim{} is very challenging in the absence of such explicit signals. 
 
The distinction between \Claim{} and \Premise{} can also be challenging, as exemplified in Figure~\ref{fig:pec_e3_p2}. 
We observe that some DMs might be associated to ADU labels more strongly than others (e.g., in Figure~\ref{fig:pec_e3_p2}, ``therefore'' is associated to \Claim{} predictions, while ``firstly'' cannot be associated to a particular label). 
Surprisingly, we observed that some DMs that are commonly associated as indicators of either \Claim{} or \Premise{} ADUs (e.g., ``because'' and ``moreover'' typically associated to \Premise{}) are not consistently used by the downstream model to predict the corresponding ADU label accordingly. 

\begin{figure}[h]
	\centerline{\includegraphics[width=0.7\textwidth,keepaspectratio]{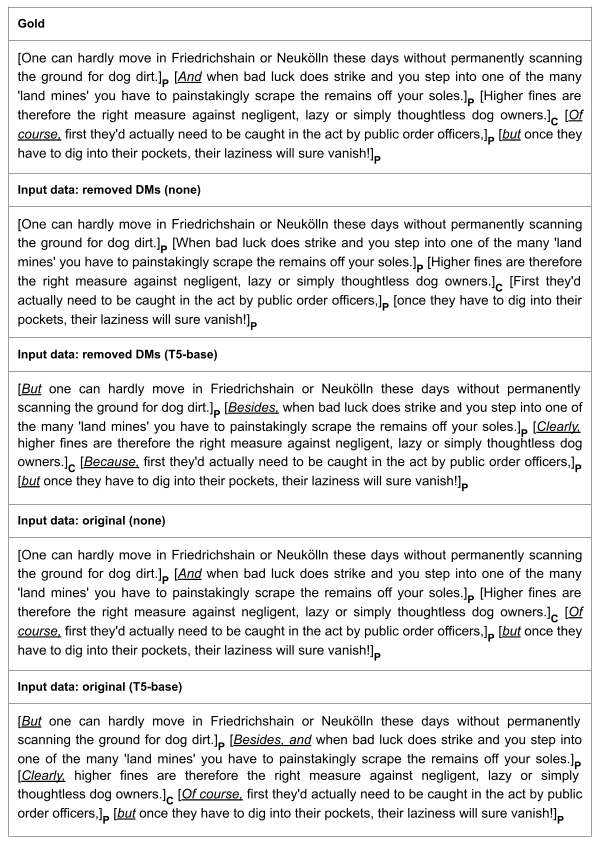}}
	\caption{MTX example.}
	\label{fig:mtx_p1}
\end{figure}

Figure~\ref{fig:mtx_p1} shows an example from MTX. 
Regarding the setups containing the original data (i.e., ``Gold'' annotations and the predictions made for ``Input data: original (none)''), besides a single occurrence of ``therefore'' and ``nevertheless'', all the remaining \Claim{} do not contain a DM preceding them (this analysis is constrained to the test set). 
Some of the ADUs labeled as \Premise{} are preceded with DMs (most common DMs are: ``and'' (6), ``but'' (10), ``yet'' (4), and ``besides'' (3)), even though most of them (44) are not preceded by a DM (numbers in parentheses correspond to the number of occurrences in the test set for the ``Gold'' annotations, similar numbers are obtained for ``Input data: original (none)''). 
DM augmentation approaches performed well in terms of coverage, with most of the ADUs being preceded by DMs. 
We can observe in Figure~\ref{fig:mtx_p1} that some ADU labels become more evident after the DM augmentation performed by the models proposed in this work (``Input data: removed DMs (\Tbase{})'' and ``Input data: original (\Tbase{})''), such as the presence of the DM ``clearly'' indicating \Claim{} and the presence of ``besides'', ``because'' or ``but'' indicating \Premise{}. 

\begin{figure}[h]
	\centerline{\includegraphics[width=0.7\textwidth,keepaspectratio]{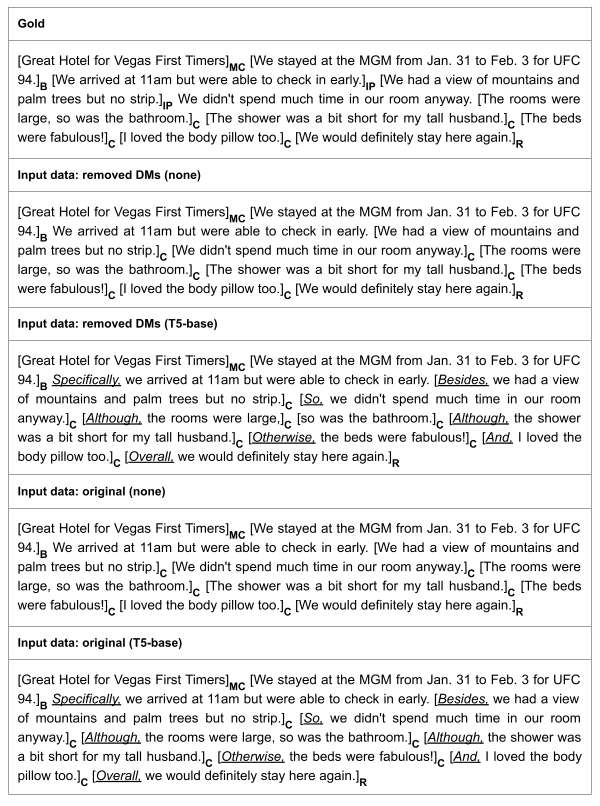}}
	\caption{Hotel example.}
	\label{fig:hotel_p5}
\end{figure}

Finally, Figure~\ref{fig:hotel_p5} shows an example from Hotel. 
Similar to the observations made for MTX, in the setups containing the original data (i.e., ``Gold'' annotations and the predictions made for ``Input data: original (none)''), most ADUs are not preceded by DMs. 
The only exception is the DM ``and'' that occurs with some frequency preceding \Claim{} (10 out of 199 ADUs labeled as \Claim{}) and \Premise{} (4 out of 41). 
For instance, in Figure~\ref{fig:hotel_p5}, 9 ADUs were annotated and none of them is preceded by a DM; making the annotation of ADUs (arguably) very challenging. 
Despite the lack of explicit clues, downstream models perform relatively well in this example, only missing the two gold \ImplicitPremise{}s (not identified as an ADU in one of the cases and predicted as \Claim{} in the other case) and erroneously labeling as \Claim{} the only sentence in the gold data that is not annotated as an ADU. 
Also similar to MTX, DM augmentation approaches performed well in terms of coverage, with most ADUs being preceded by DMs. 
However, as observed in Figure~\ref{fig:hotel_p5}, the impact on the downstream model predictions is small (the predictions for all the setups are similar, the only exception is the extra split on ``so was the bathroom'' performed in ``Input data: removed DMs (\Tbase{})'', even though this span of text is similar in all setups). 
We point out that, particularly in this text genre, adding DMs to signal the presence of ADUs might contribute to improving the readability of arguments exposed in the text, as exemplified by the DM augmentation performed by the models proposed in this work (``Input data: removed DMs (\Tbase{})'' and ``Input data: original (\Tbase{})'' in Figure~\ref{fig:hotel_p5}). 

